\pdfoutput=1
\documentclass[11pt]{article}
\usepackage{booktabs}
\usepackage{longtable}
\usepackage{amsmath}
\usepackage{booktabs}
\usepackage[preprint]{acl}

\usepackage{times}
\usepackage{latexsym}
\usepackage[utf8]{inputenc}
\usepackage[T1,T2A]{fontenc}
\usepackage[russian,english]{babel}
\usepackage{graphicx}

\usepackage{hyperref}
\usepackage{multirow}
\usepackage{adjustbox, booktabs, multirow, multicol, amssymb, url}

\usepackage[T1]{fontenc}
\usepackage[utf8]{inputenc}

\usepackage{microtype}

\usepackage{inconsolata}

\usepackage{graphicx}
\usepackage{pgfplots}
\usepgfplotslibrary{groupplots}
\pgfplotsset{compat=1.18}
\usepackage{booktabs}
\usepackage{multirow}
\usepackage[table]{xcolor}

\usepackage[many]{tcolorbox}
\usepackage{xcolor}
\usepackage{caption}
\usepackage{float}
\usepackage[utf8]{inputenc} 
\usepackage{color}
\usepackage{pifont}        % For checkmark and xmark symbols
\usepackage{xcolor}        % For coloring symbols
\usepackage{booktabs}      % For nicer table lines (optional)

\tcbuselibrary{listings}

\usepackage{pifont}

\definecolor{darkgreen}{rgb}{0.0,0.5,0.0}
\definecolor{tidered}{HTML}{C0392B}
\definecolor{tideblue}{HTML}{2980B9}
\definecolor{tidegreen}{HTML}{27AE60}
\definecolor{tideorange}{HTML}{D68910}
\definecolor{headshade}{RGB}{230,238,248}
\definecolor{groupshade}{RGB}{244,246,249}
\definecolor{cutshade}{RGB}{252,244,227}
\usepackage{colortbl}
\definecolor{tidetint}{rgb}{0.90,0.945,0.98} % light tint on the TIDE row

\usepackage{makecell}
\usepackage{tikz}
\definecolor{tideDrop}{HTML}{C0392B}
\definecolor{paraA}{HTML}{EEF0F2}   % parametric block, light ash
\definecolor{paraB}{HTML}{DDE1E6}
\definecolor{ragA}{HTML}{E1EBF7}    % RAG block, cool blue
\definecolor{ragB}{HTML}{C4D8EF}
\definecolor{iclA}{HTML}{E7F1E2}    % ICL block, soft green
\definecolor{iclB}{HTML}{CBE3C2}
\definecolor{inccol}{HTML}{0F7B6C}  % partial-credit increment, teal
\definecolor{cuecol}{HTML}{51607A}  % header cue

\newcommand{\lockshut}{\raisebox{-1pt}{\begin{tikzpicture}[x=1pt,y=1pt,line width=0.5pt,line join=round]
  \draw (-1.5,1.5) -- (-1.5,3) arc (180:0:1.5) -- (1.5,1.5);   % closed shackle
  \fill[rounded corners=0.6pt] (-2.5,-3) rectangle (2.5,1.5);  % body
  \fill[white] (0,-0.3) circle (0.5);                          % keyhole
  \fill[white] (-0.28,-1.7) rectangle (0.28,-0.3);
\end{tikzpicture}}}
\newcommand{\lockopen}{\raisebox{-1pt}{\begin{tikzpicture}[x=1pt,y=1pt,line width=0.5pt,line join=round]
  \draw (-1.5,1.5) -- (-1.5,3.4) arc (180:30:1.5);             % open shackle (gap on right)
  \fill[rounded corners=0.6pt] (-2.5,-3) rectangle (2.5,1.5);
  \fill[white] (0,-0.3) circle (0.5);
  \fill[white] (-0.28,-1.7) rectangle (0.28,-0.3);
\end{tikzpicture}}}

\definecolor{tideDrop}{HTML}{C0392B}
\definecolor{tideInc}{HTML}{2E7D32}

\tcbset{expanel/.style 2 args={enhanced, breakable,
  colback=#1!6!white, colframe=#1!50!black,
  colbacktitle=#1!50!black, coltitle=white,
  fonttitle=\bfseries, title={#2},
  boxrule=0.5pt, arc=3pt, left=5pt, right=5pt, top=4pt, bottom=4pt}}

  \definecolor{solverrose}{HTML}{C0394F}
  \definecolor{timeteal}{HTML}{23907F}
  \definecolor{genblue}{HTML}{2F6DB5}
  \definecolor{analystgreen}{HTML}{4C9136}
  \definecolor{veramber}{HTML}{B57A06}
  \definecolor{exppurple}{HTML}{6E5BC4}
  \newtcolorbox{promptbox}[1]{breakable, enhanced, colback=#1!5!white,
    colframe=#1, boxrule=0.6pt, arc=2pt, left=5pt, right=5pt, top=4pt,
    bottom=4pt, fontupper=\scriptsize, before upper={\raggedright}}
\usepackage{listings}

\lstdefinestyle{qual}{
  language=Python, basicstyle=\ttfamily\scriptsize,
  keywordstyle=\bfseries, commentstyle=\itshape,
  showstringspaces=false, upquote=true,
  columns=fullflexible, keepspaces=true, breaklines=true,
  postbreak=\mbox{\textcolor{gray}{$\hookrightarrow$}\space},
  moredelim=**[is][\color{red}]{@}{@},
  aboveskip=0pt, belowskip=0pt,
}
\tcbset{qualbox/.style={
  enhanced, breakable, colback=gray!4, colframe=gray!55,
  boxrule=0.4pt, arc=1pt, left=3pt, right=3pt, top=2pt, bottom=2pt,
  before skip=6pt, after skip=6pt,
}}

\newcommand{\UCR}{\raisebox{-0.2em}{\includegraphics[height=1em]{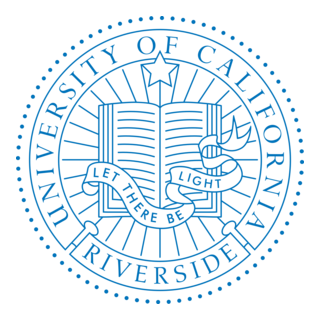}}}
\newcommand{\UIU}{\raisebox{-0.2em}{\includegraphics[height=1em]{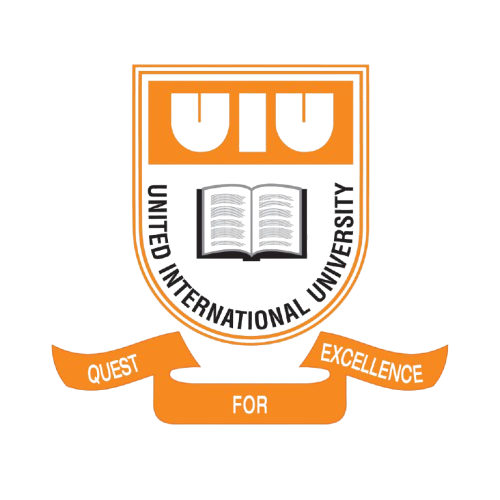}}}
\newcommand{\BRACU}{\raisebox{-0.2em}{\includegraphics[height=1em]{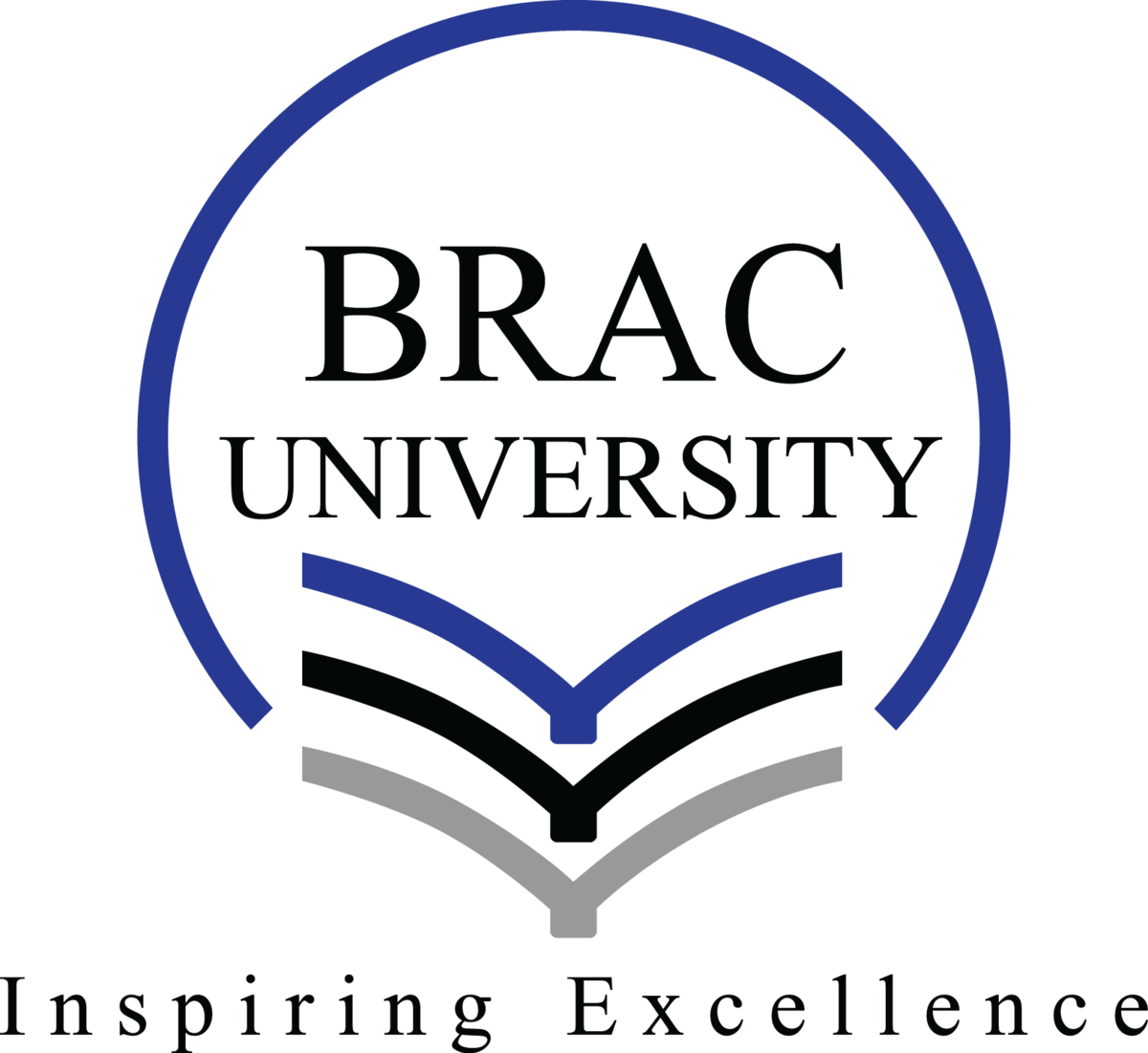}}}
\newcommand{\UMBC}{\raisebox{-0.2em}{\includegraphics[height=1em]{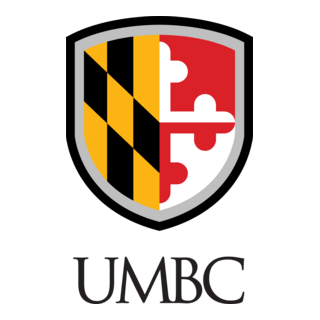}}}

\title{\textit{Did the Problem Actually Get Harder?} Testing LLMs on Paired P and NP-Complete Problems Hardened by Adversarial Multi-Agent Rewriting} 

\author{
 \textbf{Nurul Labib Sayeedi\,\UIU},
 \textbf{Md. Faiyaz Abdullah Sayeedi\,\UIU\BRACU},
 \textbf{Zabir Al Nazi\,\UCR},\\
 \textbf{Mahbub E Sobhani\,\UIU\BRACU},
 \textbf{Shubhashis Roy Dipta\,\UMBC}
\\[0.5em]
 \UIU\ United International University \quad
 \BRACU\ BRAC University
\\[0.2em]
 \UCR\ University of California, Riverside \quad
 \UMBC\ University of Maryland, Baltimore County
\\
{\small
 \texttt{\{nsayeedi2410045@bsds, msayeedi212049@bscse, msobhani2410011@mscse\}.uiu.ac.bd},} \\
{\small 
 \texttt{znazi002@ucr.edu},
 \texttt{sroydip1@umbc.edu}
}
}

\newcommand{\methodname}{\textsc{AgentGAN}}
\newcommand{\dataname}{\textsc{\textsc{P2NPCBench}}}

\begin{document}
\maketitle

\begin{abstract}
Many NP-complete (NPC) problems have a polynomial-time (P) twin that differs
in a single structural detail, such as 3- versus 2-colouring a graph. A
model that reasons rather than recalls should notice which side of this
boundary a problem falls on, yet existing benchmarks either fix the wording
of each task or ignore its complexity class, so they cannot separate the two.
We introduce \dataname{}, a benchmark of 12 problem families in which every
NPC problem is paired with a P twin in the same problem setting. To remove
the textbook cues that enable recall, we build it with \textsc{AgentGAN}, an
adversarial multi-agent framework in which a generator rewrites each
statement until a panel of frozen code models fails, while an executable
validity gate keeps every rewrite answer- and class-preserving. The result
is 2{,}114 hardened statements with human-verified reference solvers.
Evaluating 12 proprietary and open-source models, we find that every model
scores lower on NPC statements than on their P twins (35.4\% vs.\ 63.8\%
pass@1), and that 95.1\% of model--family pairs fail more often on the NPC side. A tenfold time limit raises NPC accuracy only from 28.5\% to 31.9\%.
Hardening lowers accuracy on both classes, more on P statements
(29.6 points) than on NPC statements (19.5 points). Finally, we build a failure taxonomy, where runtime errors are the largest
NPC failure type (44.1\%), and nearly half of them (46.5\%) come from
malformed or truncated code. The dataset and code are
available at \href{https://github.com/Labib1610/complexity-benchmark}{Github}.
\end{abstract}

% A tenfold time limit raises NPC accuracy only from 28.5\% to 31.9\%, so
% failures stem mainly from incorrect or non-executable solutions rather than
% from time constraints.

\section{Introduction}\label{intro}

Complexity theory separates problems solvable in polynomial time (P) from
NP-complete (NPC) problems, the hardest problems in NP (nondeterministic
polynomial time), for which no polynomial-time algorithm is known
\citep{karp1972reducibility, garey1979computers}. The boundary between the
two is often a single structural detail: 2-colouring a graph takes linear
time while 3-colouring is NP-complete. Such problems underlie
practical tasks in routing, scheduling, and network design
\citep{imajuku2026ale}, where misreading the class is costly: a solver either
runs an exhaustive search on an easy instance or, worse, applies a fast
algorithm that silently returns wrong answers on a hard one.

Large language
models (LLMs) now write competitive code, but they are mostly evaluated on
textbook-style or contest problems whose solutions may already appear in
their training data \citep{xia2025leetcodedataset, mirzadeh2025gsm}. A model
that reasons rather than recalls should choose its algorithm from the
structure of the problem, not from its name or wording. This is hard for
three reasons: memory favours textbook phrasings over real-life ones
\citep{duchnowski2025knapsack}; context can mislead, since one irrelevant
clause or misleading comment can flip the answer
\citep{mirzadeh2025gsm, lam2026codecrash, al2026dagger}; and computation stops scaling with
difficulty \citep{yang2025nondeterministic}. Testing this ability therefore
requires P and NPC problems that differ in class but not in surface form.

% ,stripped of the textbook cues that make recall possible.

Existing benchmarks cannot measure this failure. Combinatorial benchmarks such
as GraphArena \citep{tang2025grapharena} and NPPC
\citep{yang2025nondeterministic}, as well as training frameworks built on
NP-hard tasks \citep{li2025np}, treat complexity merely as a difficulty level.
Closest to our setting, NPHardEval \citep{fan2024nphardeval} spans P,
NP-complete, and NP-hard classes, but keeps a single clean wording for each
task, while EHOP \citep{duchnowski2025knapsack} rewrites problems as real-life
stories, but only for three NP-hard problems. Neither varies wording and class
together, so neither can show whether models break when a problem crosses
from P to NP-complete. Code robustness studies perturb code but not complexity
\citep{wang2023recode, lam2026codecrash}, math stress tests stay at
grade-school level \citep{mirzadeh2025gsm}, and budget studies vary tokens but
not problem class \citep{sun2025empirical, zhang2026budget}. To the best of our knowledge, no existing benchmark varies the complexity
class while holding the surface form fixed. A more detailed discussion of related work is given in Appendix~\ref{related_works}.

To address this gap, we introduce \dataname{}, a benchmark of 12 problem families
in which every NP-complete problem is paired with a polynomial-time twin posed
over the same surface, so that a single structural edit moves it across the class boundary. To remove the textbook cues
that let models recall rather than reason, we build it with
\textsc{AgentGAN}, a Generative Adversarial Network (GAN)-style framework of frozen LLM agents in which a
generator rewrites each statement until a panel of code models fails, while
an executable validity gate guarantees that every rewrite keeps its answer and
its class. Across 12 models, every model scores lower on NP-complete statements than on
their polynomial twins (35.4\% vs.\ 63.8\% pass@1), and a tenfold time limit
barely helps (28.5\% $\to$ 31.9\%), so failures stem mainly from incorrect or
non-executable solutions rather than from time constraints. \textbf{Our contributions are as follows:}
\begin{itemize}
  \item \dataname{}, to the best of our knowledge the first benchmark that pairs each
NP-complete problem with a polynomial-time variant differing in a single
structural detail, with 2{,}114 hardened statements and human-verified
reference solvers.
  \item \methodname{}, a training-free adversarial framework that hardens
  statements while keeping them answer- and class-preserving.
  \item We evaluate 12 proprietary and open-source models and show that the
  dominant failure lies in the P$\to$NPC direction, and that hardening narrows the class gap from 24.6 to 14.5 points.
\end{itemize}

\section{Methodology}\label{sec:method}

\begin{figure*}[t]
  \centering
  \includegraphics[width=\textwidth]{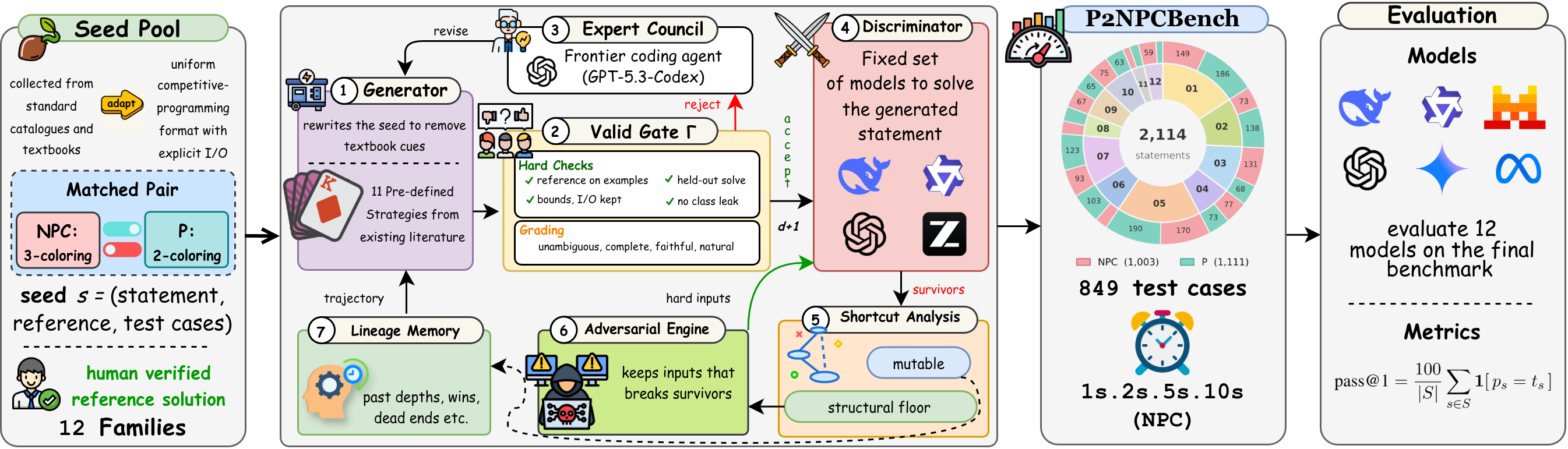}
  \caption{Overview of \methodname{}. The generator rewrites each seed,
  the validity gate $\Gamma$ admits only answer- and class-preserving drafts,
  and the frozen discriminator panel tests the accepted variants. Survivors'
  shortcuts steer the next round until every discriminator is defeated or
  $d = 10$, and the resulting statements form \dataname{}.}
  \label{fig:pipeline}
\end{figure*}

Figure~\ref{fig:pipeline} shows our overall methodology pipeline.

\subsection{Problem Formulation}\label{sec:formulation}

A \textbf{seed} $s$ is a well-posed problem with a statement, a reference
solver $R_s$, and graded tests $T_s$. Each family has two matched
\textbf{criteria} $c \in \{\textsc{npc}, \textsc{p}\}$ that pose the same task
under a controlled complexity class (e.g.\ $3$- vs.\ $2$-colouring). A
\textbf{variant} $x \in \mathcal{X}(s,c)$ is a rewrite of the seed that
preserves the ground-truth answer and the class while removing the cues that
enable textbook recall.

To measure whether a variant is harder, we use a \textbf{panel}
$\mathcal{D}=\{D_1,\dots,D_m\}$ of $m$ fixed code models, each of which
attempts to solve the variant. We write $\delta_j(x)\in\{0,1\}$ for whether
$D_j$ fails on $x$, i.e.\ its solution does not pass every test in $T_s$. A
variant is only useful if it remains valid: answer-preserving,
class-preserving, well-posed, and solvable. We enforce this with a
\textbf{validity gate} $\Gamma$, a set of checks that every variant must pass
(Section~\ref{sec:framework}), and write $\mathcal{X}_\Gamma(s,c) \subseteq
\mathcal{X}(s,c)$ for the variants it admits. Finally, let $A \subseteq
\{1,\dots,m\}$ be the indices of panel models that solve the seed itself, since
only these can still be defeated. The goal is to find the valid variant that
defeats as many of them as possible:
\begin{equation}
x^{\star}=\arg\max_{x\,\in\,\mathcal{X}_\Gamma(s,c)}\ \sum_{j\in A}\delta_j(x).
\label{eq:objective}
\end{equation}

The feasible set is defined by the validity gate $\Gamma$, so the generator
cannot win by breaking the problem: $\Gamma$ forbids altering the answer or
the class, dropping bounds or tie-break rules, leaking the class or algorithm
name, and producing an ill-posed or unsolvable statement.

\subsection{\methodname{} Framework}\label{sec:framework}

\paragraph{Adversarial roles.}
We do not propose a new multi-agent adversarial pipeline. Instead, we adapt MALLM-GAN
\citep{ling-etal-2026-mallm}, which casts frozen LLM agents as a generator and
discriminators for tabular data synthesis, to generating and hardening
\textsc{p} and \textsc{npc} problems. We name our adapted framework
\methodname{}. We also add some modules that MALLM-GAN lacks \citep{10.1145/3805712.3809931}:
a cross-round \emph{memory} that conditions the generator on the full
hardening trajectory, an executable \emph{validity gate} that prevents the
generator from winning by corrupting the problem, and an \emph{expert
council} that critiques rejected drafts. The \textbf{generator} is a single
strong model and the sole author of all statement text. The \textbf{discriminator panel} is a fixed set of models $m$. Rather than imposing one decoding configuration on the panel, each
model is run at the settings its own documentation recommends for code
generation (Appendix~\ref{app:decoding}), so that no model is penalised for a
configuration it was not tuned for and every measured failure reflects the
model at its intended operating point. The panel is
\emph{frozen} by design because the
quantity of interest is the drop induced on a fixed population, not an
equilibrium between two moving players.

% \paragraph{Validity gate.}
% $\Gamma$ combines a set of binary hard constraints with a set of
% verifier-graded criteria. A candidate is accepted only when every hard
% constraint holds \emph{and} every graded criterion clears its threshold \textcolor{red}{not clear! what is the threshold?}; the
% two are never averaged together, since a hard constraint encodes a property the
% variant must not violate at all, whereas a graded criterion encodes quality
% that admits degree \textcolor{red}{rewrite this paragraph cleanly, it's pretty confusing}. Most of the hard constraints are \emph{executable} rather
% than judged: the reference solver is run on every example embedded in the
% rewrite and must reproduce the stated output, any added formal constraint is
% evaluated against all seed inputs, and a solver round-trip requires a model
% outside the panel to solve the rewrite and pass the base test tiers. That last
% check is what keeps a variant \emph{solvable}, raising the difficulty ceiling
% without sacrificing gradeability. The remaining hard constraints are
% deterministic scans for I/O contract and bound preservation and for forbidden
% class or algorithm terms. The graded criteria cover unambiguity, completeness,
% faithfulness, and naturalness, \textcolor{red}{how? which prompt? refer to appendix?} and among accepted candidates selection
% maximises a weighted composite of them that privileges faithfulness \citep{cheng2026terminal}. The verifier is drawn from a different provider than the generator to avoid self-enhancement bias.

\paragraph{Validity gate.}
$\Gamma$ applies two kinds of checks to every candidate. \emph{Hard
constraints} are pass/fail properties that a variant must never violate, and
most of them are executable rather than judged. The reference solver is run on
every example in the rewrite and must reproduce the stated output, any added
formal constraint is evaluated against all seed inputs, and a model outside
the panel must solve the rewrite and pass the base test tiers, which ensures
that the variant remains solvable. Deterministic scans further check that the
I/O contract and all bounds are preserved and that no class or algorithm name
appears. \emph{Graded criteria} measure quality: a verifier from a different
provider than the generator, chosen to avoid self-enhancement bias, scores
unambiguity (G1), completeness (G2), faithfulness (G3), and naturalness (G4)
on a 0--5 scale, using the prompt in Appendix~\ref{app:prompts}. A candidate
is accepted only if it passes every hard constraint and reaches the minimum
score on every criterion. The two kinds of checks are applied separately rather than
combined into one score, so a high quality score cannot make up for a violated
constraint. Among accepted candidates, we keep the one with the highest
weighted sum of the four scores, with faithfulness weighted most
\citep{cheng2026terminal}.

\noindent \textbf{Hardening loop.}
Each accepted base variant seeds a \textbf{lineage} hardened over depths
$d=0,\dots,d_{\max}$ with $d_{\max}{=}10$. At depth $d$ the panel is run on
$x_d$; the verifier summarises the single shortcut the survivors exploited and
labels it \emph{mutable} (rewordable) or a \textbf{structural floor} (the
shortcut \emph{is} the task). The generator then produces candidates along a
fallback ladder, each gate-validated, and the highest-$Q$ accepted candidate
becomes $x_{d+1}$. Defeated discriminators are \emph{retired}, still
re-tested at each depth to confirm continued defeat, but no longer steering the
generator:
\begin{equation}
A_{d+1}=A_d \setminus \{\,j : \delta_j(x_d)=1\,\}
\label{eq:loop}
\end{equation}
We report $d^{\star}=\min\{d : A_{d+1}=\emptyset\}$, the
\textbf{depth-to-defeat}, as the per-lineage hardness statistic; a shallow
$d^{\star}$ indicates a seed that was already hard.

\paragraph{Memory and escalation.}
A per-lineage memory $M$ records the depth history, which strategies newly
defeated or regressed which discriminators, dead ends, and per-model profiles.
A token-capped block is fed back into $G$, so the generator conditions on the
full trajectory rather than the last round alone, the closest analogue to
gradient accumulation available without weight updates. Two escalations fire
when the ladder stalls: \emph{aggressive mode} ($d\ge3$) requires the generator
to invent and name a novel strategy, persisted to the library; and the
\emph{adversarial corner engine} pivots from surface attack to \textbf{instance
attack} once a structural floor is declared, generating a candidate input pool
and retaining those that break survivors under differential testing
against $R_s$.

\paragraph{Expert council.}
When $\Gamma$ rejects a draft, an expert model from a third provider critiques
it and proposes alternatives, and the generator revises over bounded
negotiation rounds with the transcript persisted. The expert never authors
final text and never validates; role separation (generator, verifier, panel, expert, held-out) is enforced in code, and no model may
occupy two roles.

\subsection{\dataname{} Dataset}\label{sec:dataset}

\begin{table}[!t]
\centering
\begin{adjustbox}{width=\columnwidth}
\small
\setlength{\tabcolsep}{3pt}
\begin{tabular}{llll}
\toprule
\rowcolor{tidetint} \textbf{\#} & \textbf{Family} & \textbf{\textsc{npc} criterion} & \textbf{\textsc{p} relative} \\
\midrule
01 & Graph colouring       & $3$-colouring              & $2$-colouring \\
02 & Satisfiability        & $3$-SAT                    & $2$-SAT \\
03 & Matching              & $3$-D matching             & Bipartite matching \\
04 & Clique / ind.\ set    & General graphs             & Interval graphs \\
05 & Steiner tree          & Terminal subset            & Spanning tree (MST) \\
06 & Job-shop              & Job-shop makespan          & Two-machine flow shop \\
07 & Exact cover           & Exact cover by $3$-sets    & Exact cover by $2$-sets \\
08 & Linear arrangement    & General graphs             & Trees \\
09 & Multicommodity flow   & $k$-commodity (integer)    & Single commodity \\
10 & Edge colouring        & Cubic graphs               & Bipartite graphs \\
11 & Multiway cut          & $\geq 3$ terminals         & $s$--$t$ min cut \\
12 & Precedence scheduling & $\geq 2$ machines          & Single machine \\
\midrule
\end{tabular}
\end{adjustbox}
\caption{The twelve families of \dataname{}, each pairing an \textsc{npc} criterion
with a polynomial-time relative one structural edit away.}
\label{tab:families}
\end{table}

\paragraph{Provenance of the seeds.}
Seed problems were not authored by us. Each family's \textsc{npc} criterion is
drawn from the standard catalogue of \textsc{np}-complete problems: the
canonical list of \citet{garey2002computers}, the \textsc{np}-completeness
chapters of widely used algorithms textbooks
\citep{cormen1990introduction,kleinberg2006algorithm}, and two openly
maintained online catalogues: the compendium of \textsc{np} optimization
problems\footnote{\url{https://www.csc.kth.se/tcs/compendium/}} and the
community-maintained list of \textsc{np}-complete
problems.\footnote{\url{https://en.wikipedia.org/wiki/List_of_NP-complete_problems}}
Problem statements and candidate reference solvers were collected from these
sources and then adapted into a uniform competitive-programming format with an
explicit I/O contract, numeric bounds, and tie-break rules. This yields 24
reference solvers, one for each criterion of the 12 families (12 \textsc{npc}
and 12 \textsc{p}). Every reference
solver was subsequently run by the authors on the full test suite of its
family and checked by hand against the source's stated semantics. Solvers that
disagreed with the source, mishandled boundary cases, or left the tie-break
behaviour underdetermined were rewritten until they did not. All authors hold
computer science degrees and are regular ICPC regional contestants, so the verification was carried out by annotators already fluent in
both the complexity-theoretic content and the competitive-programming
conventions the format assumes. Every reference answer was independently checked by two authors, with a high
inter-annotator agreement of Cohen's $\kappa = 0.86$. This manual pass matters because the entire
pipeline treats $R_s$ as ground truth.

% \paragraph{Family inclusion criterion.}
% A family enters \dataname{} only if it admits a \textbf{matched pair}: an
% \textsc{npc} criterion and a polynomial-time relative that can be posed over
% the \emph{same surface}, differing in the single structural feature that moves
% the problem across the class boundary. This is a strong filter, and it is what
% makes the bidirectional comparison well defined --- without a same-surface
% polynomial relative there is no $\textsc{p}\!\rightarrow\!\textsc{npc}$
% direction to measure, and any observed accuracy gap would confound class with
% domain. Twelve families satisfy it, as listed in Table~\ref{tab:families}.

% \paragraph{Excluded families.}
% Some candidate families were built and then dropped because no suitable
% relative exists. For example, \textbf{Supply Purchasing} (Quadratic Diophantine
% Equations) is \textsc{np}-complete, but no polynomial variant is reachable by a
% structural edit that preserves the surface: relaxing it changes the task rather
% than its class. \textbf{Network Broadcast Design} (Maximum Leaf Spanning Tree)
% becomes easy on trees, but this easy version was not suitable for the benchmark: a
% tree is its own only spanning tree, so the task reduces to counting leaves and
% leaves no polynomial algorithm on the \textsc{p} side to contrast with the
% \textsc{npc} one.

\paragraph{Family inclusion criterion.}
A family is included in \dataname{} only if it admits a \textbf{matched pair}: an
\textsc{npc} criterion and a polynomial-time relative posed in the
\emph{same problem setting} and differing in the single structural feature
that moves the problem across the class boundary. Without such a relative
there is no $\textsc{p}\!\rightarrow\!\textsc{npc}$ direction to measure, and
any accuracy gap would confound class with domain. Twelve families satisfy this criterion and are listed in Table~\ref{tab:families}.

\paragraph{Excluded families.}
We dropped candidate families for one of two reasons. First, the family has no
polynomial variant that keeps the same problem setting: for
\textbf{Supply Purchasing} (Quadratic Diophantine Equations), any relaxation
changes the task rather than its class. Second, the polynomial variant is too
easy to contrast with the \textsc{npc} one: \textbf{Network Broadcast Design}
(Maximum Leaf Spanning Tree) becomes polynomial on trees, but a tree is its own
only spanning tree, so the task reduces to counting leaves.

\paragraph{Class-disguise edits.}
Eleven strategies $\sigma\in\{S_{01},\dots,S_{11}\}$ are applied at one of
three intensities: \emph{light}, \emph{moderate}, or \emph{aggressive},
which control how far the rewrite departs from the seed, from a minimal
surface change to a full re-authoring of the statement, and, where compatible,
composed in pairs at the two higher intensities. The strategies are:
$S_{01}$ removes textbook-recall wording \citep{wang-etal-2023-recode};
$S_{02}$ keeps the structure but changes the real-world scenario \citep{duchnowski-etal-2025-knapsack, ICLR2025_ec2e7a89};
$S_{03}$ describes special structure by its mechanics, never by name \citep{wang-etal-2023-recode};
$S_{04}$ makes a hard problem look as though it admits an easy shortcut \citep{NEURIPS2025_aec5e284};
$S_{05}$ makes an easy problem look as though it demands heavy machinery \citep{NEURIPS2025_aec5e284};
$S_{06}$ states bounds indirectly, so the reader must derive them \citep{wang-etal-2023-recode};
$S_{07}$ embeds the task in a scenario where every detail is load-bearing \citep{ICLR2025_ec2e7a89};
$S_{08}$ adds true non-binding, answer-preserving constraints \citep{ICLR2025_ec2e7a89};
$S_{09}$ presents the task as one step in a larger workflow \citep{ICLR2025_ec2e7a89};
$S_{10}$ reorders the statement so the objective arrives after the constraints \citep{wang-etal-2023-recode};
and $S_{11}$ attacks the specific shortcut observed in discriminator reasoning
traces, in a primary and a secondary form. 

% $S_{04}$ and $S_{05}$ are
% directional and carry the central manipulation; $S_{11}$ is the hardening
% loop's primary lever.

% Curves are statement-weighted mixtures of Beta-PERT densities fitted to each family's mean and min--max length.

\paragraph{Test case tiers.}
Every criterion is graded on three separately scored tiers: \textbf{anchor}
(small seed cases), \textbf{corner} (generated adversarial edge cases), and
\textbf{worst-case} (large or near-threshold instances). Grading is either
exact match against $R_s$ or witness verification via a per-family validator.

\paragraph{Human verification of hardened statements.}
The validity gate is automated, so we also checked its output by hand. From
each of the 12 families we randomly sampled 20 hardened statements per
criterion, giving 240 \textsc{npc} and 240 \textsc{p} statements (480 in
total). Each was shown alongside its seed to three undergraduate annotators,
all rated within 1200--1399 (Pupil) on Codeforces, who scored it on a
three-point scale: \textbf{0} if the rewrite is not faithful to the seed,
i.e.\ it changes the task or introduces an error; \textbf{1} if it is faithful
but stays close to the seed's wording; and \textbf{2} if it departs
substantially from the seed while remaining correct and faithful. The scale
thus measures both properties: a 0 marks a
failure of the gate, and the split between 1 and 2 shows how much disguise the
hardening achieved. By majority vote, annotators assigned a score of 0 to
4.2\% of statements (20 of 480) with inter-annotator agreement of Fleiss'
$\kappa=0.79$. The 20 statements scored 0 were removed from the benchmark;
the full breakdown is given in Appendix~\ref{app:human_verification}.

\subsection{Dataset Statistics}\label{sec:stats}

\begin{figure}[t]
  \centering
  \includegraphics[width=0.9\columnwidth]{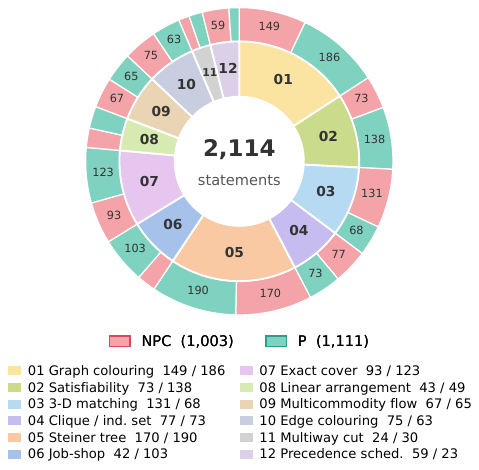}
  \caption{Composition of \dataname{}. Inner ring: families sized by evaluated statements.
  Outer ring: NPC/P split within each family.}
  \label{fig:composition}
\end{figure}

\begin{figure}[t]
  \centering
  \includegraphics[width=0.8\columnwidth]{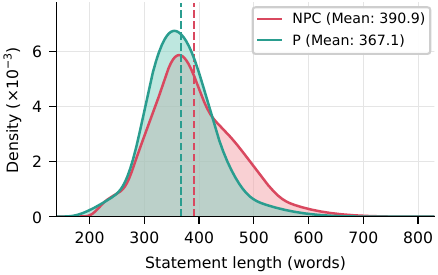}
  \caption{Length distribution of the statements.}
  \label{fig:length_classes}
\end{figure}

\dataname{} comprises 12 families, each with an \textsc{npc} and a \textsc{p}
criterion, totalling 2{,}114 evaluated statements (1{,}003 \textsc{npc},
1{,}111 \textsc{p}) drawn from 1{,}141 accepted versions; a statement counts every version together with every depth of its lineage, so statements within a lineage are related rather than independent
(Figure~\ref{fig:composition}). The two classes are closely balanced in every
structural dimension (580 vs.\ 561 versions, 152 vs.\ 143 lineages, and
434 vs.\ 415 test cases) so neither direction of the
$\textsc{p}\!\leftrightarrow\!\textsc{npc}$ comparison rests on a larger
sample. Statements average 378.4 words (median 370,
s.d.\ 78.9, range 161--804), well above typical competitive-programming
problems by design: hardened statements embed load-bearing context that the
model must read to recover the task. \textsc{npc} statements are slightly
longer than their \textsc{p} counterparts (390.9 vs.\ 367.1 words on average;
Figure~\ref{fig:length_classes}).

% Most families reach the full complement of 14 lineages per criterion;
% linear arrangement, multiway cut, and the \textsc{p} side of precedence
% scheduling stop earlier, at 3--9.

\section{Experimental Setup}\label{setup}

\subsection{Models}
The hardening pipeline uses Gemini~3.1~Pro as the \textbf{generator};
DeepSeek~V4~Flash \citep{deepseekai2026deepseekv4highlyefficientmilliontoken}, GPT-OSS-120B \citep{openai2025gptoss120bgptoss20bmodel}, Qwen3~Coder \cite{cao2026qwen3codernexttechnicalreport}, and GLM-5 \citep{glm5team2026glm5vibecodingagentic} as the
\textbf{discriminator panel}; GPT-5.6~Luna as the \textbf{verifier}; and
GPT-5.3-Codex as the \textbf{expert}. Kimi~K3 \citep{kimiteam2026kimik3openfrontier} is a \ textbf {held-out} model: unseen by the pipeline and used only for final evaluation. We evaluate twelve models on the final
benchmark: GPT-5.6~Sol, Gemini (3.8-Flash, 3.7-Flash), Qwen3.8 (Flash \citep{qwen3.8flashnext},
Omni-Flash \citep{qwenteam2026qwen38omninativeomnimodalagents}), Mistral-Medium-3.5, DeepSeek-V4.1-Flash \citep{xu2026deepseek}, Gemma-3 (4B, 12B, 27B) \citep{gemmateam2025gemma3technicalreport},
Qwen3.8-27B \citep{qwen38}, and Llama-4-Scout. The decoding settings
below apply to these twelve benchmark models. The models in the hardening
pipeline use the settings listed in Appendix~\ref{app:decoding}. Decoding is
identical across the benchmark models so that differences are attributable to
the model rather than to sampling: temperature $0.0$, top-$p$ $1.0$, and a
maximum of 32{,}768 output tokens. Each model produces a single Python~3
solution per statement.

\begin{table*}[!t]
\centering
\begin{adjustbox}{width=\textwidth}
\scriptsize
\setlength{\tabcolsep}{2.4pt}
\newcommand{\grouphead}[1]{\multicolumn{27}{l}{\textit{#1}}}
\begin{tabular}{l*{13}{r}@{\hspace{8pt}}*{13}{r}}
\toprule
 & \multicolumn{13}{c}{\textbf{\textsc{npc} statements}} & \multicolumn{13}{c}{\textbf{\textsc{p} statements}} \\
\cmidrule(lr){2-14}\cmidrule(lr){15-27}
\textbf{Model} & 1 & 2 & 3 & 4 & 5 & 6 & 7 & 8 & 9 & 10 & 11 & 12 & \textbf{Avg} & 1 & 2 & 3 & 4 & 5 & 6 & 7 & 8 & 9 & 10 & 11 & 12 & \textbf{Avg} \\
\midrule
\grouphead{\lockshut~Proprietary models} \\
GPT-5.6 Sol & \underline{52.4} & \underline{76.9} & 78.4 & \underline{75.4} & \underline{76.9} & \cellcolor{iclA}\underline{87.4} & \underline{46.4} & \textbf{67.5} & 54.4 & \textbf{82.7} & \underline{49.5} & \textbf{73.2} & \cellcolor{tidetint}\textbf{68.4} & \cellcolor{iclA}\textbf{100.0} & \textbf{97.4} & \textbf{92.3} & \textbf{98.5} & 95.2 & \cellcolor{iclA}\textbf{100.0} & \underline{95.1} & \textbf{82.1} & \underline{95.6} & \textbf{86.2} & \underline{93.7} & 96.8 & \cellcolor{tidetint}\textbf{94.4} \\
Gemini-3.8-Flash & \textbf{55.1} & 30.1 & \textbf{82.3} & 73.1 & 65.9 & \cellcolor{iclA}\textbf{89.3} & \textbf{47.3} & \underline{46.1} & \textbf{77.1} & 62.7 & \textbf{53.3} & \underline{69.5} & \cellcolor{tidetint}\underline{62.6} & \underline{99.5} & 81.2 & 86.2 & \textbf{98.5} & \textbf{99.6} & \cellcolor{iclA}\textbf{100.0} & 92.3 & 72.9 & \cellcolor{iclA}\textbf{100.0} & 55.4 & \cellcolor{iclA}\textbf{100.0} & \cellcolor{iclA}\textbf{100.0} & \cellcolor{tidetint}\underline{90.5} \\
Gemini-3.7-Flash & 49.5 & 48.3 & 64.7 & 66.3 & \cellcolor{iclA}\textbf{77.3} & 72.6 & 32.6 & 42.3 & \underline{74.7} & \underline{63.2} & 39.1 & 63.4 & \cellcolor{tidetint}57.8 & \cellcolor{iclA}\textbf{100.0} & \underline{88.3} & \underline{87.6} & 95.2 & 93.8 & \cellcolor{iclA}\textbf{100.0} & 92.8 & \underline{81.2} & 92.8 & \underline{66.4} & 92.3 & 93.5 & \cellcolor{tidetint}90.3 \\
\midrule
\grouphead{\lockopen~Open-source models} \\
Qwen3.8-Flash & 46.8 & 61.6 & \cellcolor{iclA}\underline{81.6} & 39.0 & 38.5 & 13.1 & 21.5 & 2.3 & 61.2 & 30.7 & 41.7 & 35.6 & \cellcolor{tidetint}39.5 & 85.1 & 76.8 & 82.3 & 93.6 & 87.2 & \cellcolor{iclA}\underline{94.2} & 88.8 & 64.7 & 71.3 & 39.2 & 93.3 & 91.4 & \cellcolor{tidetint}80.7 \\
DeepSeek-V4.1-Flash & 51.3 & 38.4 & \cellcolor{iclA}63.4 & 35.1 & 45.3 & 33.3 & 31.2 & 23.5 & 51.5 & 18.7 & 28.3 & 51.8 & \cellcolor{tidetint}39.3 & 93.0 & 87.0 & 83.6 & \underline{96.5} & 89.2 & 82.8 & \cellcolor{iclA}\textbf{97.0} & 62.8 & 60.6 & 14.1 & 62.6 & 87.3 & \cellcolor{tidetint}76.4 \\
Qwen3.8-Omni-Flash & 41.3 & 32.9 & \cellcolor{iclA}74.8 & 48.1 & 26.5 & 21.4 & 18.3 & 9.3 & 45.5 & 9.3 & 29.2 & 30.5 & \cellcolor{tidetint}32.3 & 90.3 & 68.1 & 76.3 & 94.7 & \cellcolor{iclA}\underline{95.8} & 92.9 & 76.1 & 52.7 & 62.7 & 11.2 & 66.7 & 85.7 & \cellcolor{tidetint}72.8 \\
Gemma-3-12B & 31.7 & \cellcolor{iclA}71.4 & 65.6 & 59.7 & 7.3 & 3.1 & 10.3 & 9.4 & 5.7 & 6.2 & 14.5 & 31.7 & \cellcolor{tidetint}26.4 & 40.1 & 73.2 & 63.1 & 64.0 & 39.2 & 67.7 & 17.0 & 13.3 & 10.2 & 11.4 & 24.7 & \cellcolor{iclA}\underline{98.4} & \cellcolor{tidetint}43.5 \\
Gemma-3-27B & 34.2 & \cellcolor{iclA}\textbf{78.7} & 73.5 & 22.1 & 12.4 & 6.0 & 9.1 & 11.2 & 14.6 & 4.8 & 9.3 & 24.2 & \cellcolor{tidetint}25.0 & 30.4 & \cellcolor{iclA}87.7 & 33.2 & 69.6 & 38.8 & 74.1 & 27.9 & 12.8 & 22.2 & 5.8 & 13.3 & 82.8 & \cellcolor{tidetint}41.5 \\
Mistral-Medium-3.5 & 32.8 & 38.4 & 71.8 & \cellcolor{iclA}\textbf{76.2} & 3.5 & 8.3 & 5.4 & 14.2 & 3.0 & 3.8 & 12.5 & 6.8 & \cellcolor{tidetint}23.1 & 89.7 & 42.0 & 74.4 & \cellcolor{iclA}94.0 & 50.0 & 38.0 & 52.1 & 22.0 & 46.2 & 12.1 & 31.8 & 61.3 & \cellcolor{tidetint}51.1 \\
Llama-4-Scout & 29.6 & 39.7 & 46.6 & \cellcolor{iclA}72.7 & 7.6 & 6.0 & 8.7 & 9.2 & 8.5 & 4.9 & 9.5 & 14.4 & \cellcolor{tidetint}21.4 & 55.1 & 32.4 & 51.9 & \cellcolor{iclA}73.7 & 31.0 & 32.0 & 9.3 & 12.8 & 12.5 & 10.4 & 16.7 & 52.4 & \cellcolor{tidetint}32.5 \\
Qwen3.8-27B & 26.4 & 24.7 & \cellcolor{iclA}67.2 & 24.7 & 26.8 & 1.2 & 17.7 & 2.3 & 6.0 & 1.3 & 18.8 & 37.3 & \cellcolor{tidetint}21.2 & 87.6 & 80.4 & 84.3 & \cellcolor{iclA}91.7 & 90.7 & 80.1 & 91.2 & 63.0 & 70.6 & 4.7 & 76.0 & 85.2 & \cellcolor{tidetint}75.5 \\
Gemma-3-4B & \cellcolor{iclA}28.3 & 4.7 & 2.8 & 23.4 & 5.8 & 1.6 & 3.8 & 4.3 & 2.3 & 1.6 & 5.2 & 3.6 & \cellcolor{tidetint}7.3 & 25.2 & 34.6 & 20.1 & \cellcolor{iclA}42.2 & 9.0 & 21.1 & 7.4 & 6.1 & 7.3 & 3.7 & 11.4 & 8.7 & \cellcolor{tidetint}16.4 \\
\midrule
\rowcolor{tidetint}
\textit{Family avg} & 40.0 & 45.5 & 64.4 & 51.3 & 32.8 & 28.6 & 21.0 & 20.1 & 33.7 & 24.2 & 25.9 & 36.8 & \textbf{35.4} & 74.7 & 70.8 & 69.6 & 84.3 & 68.3 & 73.6 & 62.2 & 45.5 & 54.3 & 26.7 & 56.9 & 78.6 & \textbf{63.8} \\
\bottomrule
\end{tabular}
\end{adjustbox}
\caption{pass@1 (\%) on \dataname{} by family; columns 1--12 follow the
numbering of Table~\ref{tab:families}, and \textbf{Avg} is the macro-average
over families. The last row averages each column over all twelve models.
Models are grouped by access and ordered by \textsc{npc}
average within each group. Within each column, \textbf{bold} marks the best
and \underline{underline} the second-best model (ties share a mark);
\colorbox{iclA}{shading} marks each model's strongest family within each
class.}
\label{tab:bench_results}
\end{table*}

\subsection{Metrics}
A solution is run on each test case of its statement and passes a case only if
it exits cleanly, finishes within the wall-clock limit, and its output is
accepted by the family's grader: a witness validator that checks the answer
against the problem constraints, or exact match against the reference output
for optimisation families. Our main metric is \textbf{pass@1}: each model
produces a single solution per statement, and the statement counts as solved
only if that solution passes \emph{every} test case; a single failed case
fails the statement:
\begin{equation}
\text{pass@1} = \frac{100}{|S|} \sum_{s \in S} \mathbf{1}\!\left[\,p_s = t_s\,\right],
\label{eq:pass1}
\end{equation}
where $p_s$ and $t_s$ are the passed and total test cases of statement $s$,
and $S$ is the set of statements a model attempted. \textsc{npc} statements are scored under four time limits
(1, 2, 5, 10\,s), which the model is told. A case passes at limit $L$ if it is
correct and its measured runtime is at most $L$. \textsc{p} statements are
scored once, at the limit stated in the problem (2s).

% As a finer-grained
% secondary measure, we also report the case-level pass rate
% $100 \times \sum_s p_s / \sum_s t_s$, which credits partially correct
% solutions.

\section{Results and Analysis}\label{results}
\subsection{Main Results}

\subsubsection{Discriminator Results}\label{sec:disc_results}

Table~\ref{tab:disc_results} reports the pass@1\% of each discriminator on the
final hardened statements. After hardening, a large and consistent gap
separates the two classes: averaged over families, accuracy on \textsc{npc}
statements is 41.9\% against 70.7\% on their \textsc{p} counterparts, and every
discriminator falls by 25--34 points when moving from \textsc{p} to
\textsc{npc}, from GPT-OSS-120B (80.1 $\rightarrow$ 45.7) to GLM-5, the weakest
on both classes (60.9 $\rightarrow$ 35.1). \textsc{npc} accuracy is lower in
eleven of twelve families, and the gap exceeds 40 points in five. The two
outliers are informative rather than contradictory: on multicommodity flow,
three of four models score higher on \textsc{npc} than on \textsc{p}, and on
edge colouring both classes sit near the floor (13.9\% and 17.4\%), so no
class gap can emerge. The gap also transfers beyond the panel: the held-out
Kimi K3, which never interacted with the pipeline during hardening, scores
87\% on \textsc{p} but only 37\% on \textsc{npc}, a 50-point gap wider than
that of any discriminator.

\begin{table}[!t]
\centering
\begin{adjustbox}{width=\columnwidth}
\footnotesize
\setlength{\tabcolsep}{3pt}
\begin{tabular}{l cccc cccc}
\toprule
 & \multicolumn{4}{c}{\textbf{\textsc{npc} (\%)}} & \multicolumn{4}{c}{\textbf{\textsc{p} (\%)}} \\
\cmidrule(lr){2-5}\cmidrule(lr){6-9}
\textbf{Family} & \textbf{OSS} & \textbf{DS} & \textbf{Qwen} & \textbf{GLM}
                & \textbf{OSS} & \textbf{DS} & \textbf{Qwen} & \textbf{GLM} \\
\midrule
Graph colouring     & 61.9 & 49.2 & 43.3 & 36.2 & 96.9 & 100  & 93.5 & 98.5 \\
SAT                 & 45.3 & 41.7 & 41.5 & 29.2 & 93.1 & 87.2 & 81.5 & 75.3 \\
3-D matching        & 62.3 & 55.4 & 48.3 & 40.7 & 87.2 & 82.6 & 76.5 & 74.9 \\
Clique / ind.\ set  & 47.4 & 43.5 & 48.6 & 32.3 & 96.2 & 87.1 & 89.7 & 70.5 \\
Steiner tree        & 75.8 & 73.6 & 64.7 & 68.2 & 92.4 & 72.1 & 82.6 & 73.3 \\
Job-shop            & 82.7 & 70.4 & 68.3 & 54.6 & 100  & 86.1 & 91.5 & 88.2 \\
Exact cover         & 26.3 & 27.1 & 23.2 & 16.6 & 88.5 & 72.4 & 45.8 & 52.7 \\
Linear arrangement  & 19.7 & 22.3 & 21.8 & 18.9 & 32.1 & 63.4 & 56.6 & 34.2 \\
Multicommodity flow & 38.1 & 58.6 & 75.9 & 68.2 & 79.3 & 46.7 & 49.6 & 52.4 \\
Edge colouring      & 18.2 & 14.7 & 16.5 & 6.4  & 17.3 & 15.8 & 21.6 & 15.1 \\
Multiway cut        & 23.7 & 11.9 & 26.4 & 12.2 & 87.4 & 83.3 & 47.6 & 17.1 \\
Precedence sched.   & 47.3 & 52.7 & 40.4 & 37.8 & 90.3 & 87.7 & 81.3 & 78.1 \\
\midrule
\rowcolor{tidetint}
\textbf{Macro-Avg}       & \textbf{45.7} & \textbf{43.4} & \textbf{43.2} & \textbf{35.1}
                    & \textbf{80.1} & \textbf{73.7} & \textbf{68.2} & \textbf{60.9} \\
\bottomrule
\end{tabular}
\end{adjustbox}
\caption{pass@1\% of the discriminators.
\textbf{OSS}: GPT-OSS-120B, \textbf{DS}: DeepSeek~V4~Flash,
\textbf{Qwen}: Qwen3~Coder, \textbf{GLM}: GLM-5.}
\label{tab:disc_results}
\end{table}

\begin{figure*}[!t]
\centering
\includegraphics[width=0.32\textwidth]{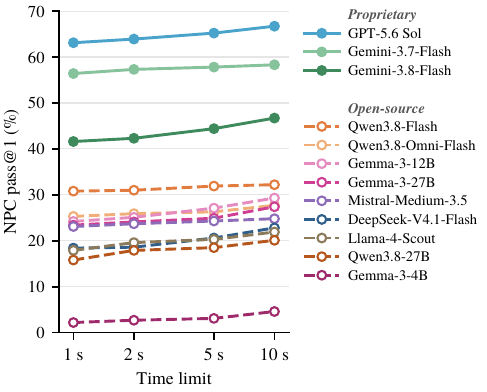}\hfill
\includegraphics[width=0.32\textwidth]{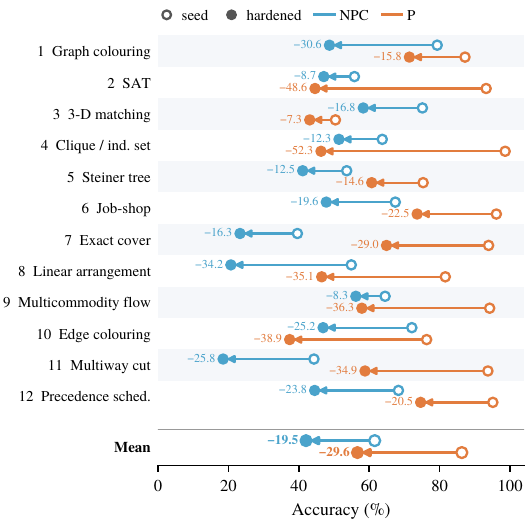}\hfill
\includegraphics[width=0.32\textwidth]{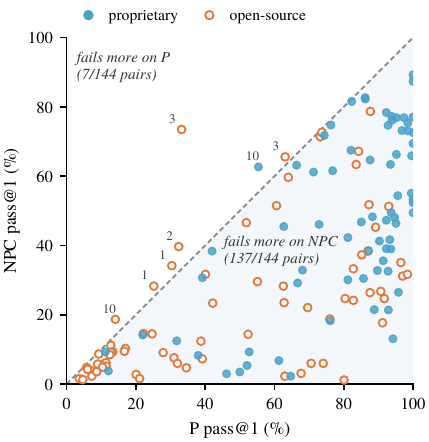}
\caption{\textbf{Left:} \textsc{npc} pass@1 under time limits. Solid
lines are proprietary and dashed lines are open-source models.
\textbf{Center:} Accuracy on the seed and on its final hardened
variants for each family, with the drop in points. \textbf{Right:}
\textsc{npc} versus \textsc{p} pass@1 for each of the 144 model--family
pairs (12 models $\times$ 12 families). Points below the diagonal fail
more often on the \textsc{npc} criterion.}
\label{fig:combined}
\end{figure*}

\subsubsection{Benchmarking Results}\label{sec:bench_results}

\noindent\textbf{Overall results.}
No model, regardless of size or access, solves \textsc{npc} statements as
reliably as their \textsc{p} counterparts. Table~\ref{tab:bench_results}
reports pass@1 for twelve models on \dataname{}. Every model scores lower on
\textsc{npc} than on \textsc{p} statements on average (35.4\% vs.\ 63.8\%
across models), with gaps from 9 to 54 points; proprietary models average
62.9\% / 91.7\% against 26.2\% / 54.5\% for open-source ones. The three
proprietary models lead on both classes: GPT-5.6~Sol (68.4\% / 94.4\%) and
the two Gemini Flash models win ten of twelve \textsc{npc} families, and the
exceptions, SAT (2) and clique (4), are won by the open-source Gemma-3-27B and
Mistral-Medium-3.5, but none closes the gap. The sharpest split is
Qwen3.8-27B, which is competitive on \textsc{p} (75.5\%) yet collapses on
\textsc{npc} (21.2\%). Small gaps among the weakest models (e.g.\ Gemma-3-4B,
7.3\% / 16.4\%) reflect failure on both classes rather than class robustness.
Linear arrangement (8) and exact cover (7) are the hardest \textsc{npc}
families, and edge colouring (10) remains hard on both sides for all but the
strongest three models. Averaged over models (last row), the class gap is
widest on job-shop (6; 28.6\% vs.\ 73.6\%), precedence scheduling (12) and
exact cover (7), each above 40 points, and in eight of twelve families every
model scores lower on \textsc{npc} than on \textsc{p}. Beyond edge colouring,
the gap nearly vanishes only on 3-D matching (3; 64.4\% vs.\ 69.6\%), whose
\textsc{npc} statements are the easiest in the benchmark, suggesting that its
hardened variants still leave a recognisable shortcut.

\noindent\textbf{Scaling Effect.}
Within the Gemma-3 family, scale helps only up to a point and never closes the
class gap. Moving from 4B to 12B lifts accuracy on both classes (7.3\% $\to$
26.4\% on \textsc{npc}, 16.4\% $\to$ 43.5\% on \textsc{p}), but 27B brings no
further gain.

% Because \textsc{p} accuracy rises faster than \textsc{npc} accuracy, the gap nearly doubles from 4B to 12B (9.1 $\to$ 17.1 points) and stays there at 27B (16.5), so the larger models do not become any better at handling the NP-complete side relative to the polynomial one.

\subsection{Effect of the Time Limit}\label{sec:time_limit}

% \begin{figure}[!t]
% \centering
% \includegraphics[width=0.9\columnwidth]{figures/time_limit.pdf}
% \caption{\textsc{npc} pass@1 under time limits. Solid lines are
% proprietary and dashed lines open-source models.}
% \label{fig:time_limit}
% \end{figure}

% \textsc{npc} statements are scored under four time limits,
% which the model is told in advance. Figure~\ref{fig:combined} \textbf{(Left)}
% shows that extra time helps little: raising the limit tenfold lifts mean
% pass@1 only from 28.5\% to 31.9\%, and no model gains more than 5.1 points,
% with GPT-5.6~Sol, for instance, moving from 63.1\% to 66.7\%. A case that
% passes only under a longer limit is correct but slow, so the small gain means
% that most failures are wrong or crashing programs rather than slow ones: the
% hardened statements defeat models by leading them to an incorrect solution,
% not by out-scaling a correct one. The gain is not front-loaded \textcolor{red}{confusing word, use something simpler}, as 45\% of it
% arrives only between 5 and 10\,s, consistent with a minority of correct
% exhaustive solvers that finish just past the shorter limits. Proprietary and
% open-source models gain about equally (3.5 vs.\ 3.3 points), and the ranking
% is almost unchanged across limits: between 1\,s and 10\,s only Gemma-3-12B
% overtakes Qwen3.8-Omni-Flash, and the gap of about 34 points between the two
% groups persists at every limit. A more generous time budget therefore does
% not alter the conclusions drawn at any single limit \textcolor{red}{ drawn at any single limit - isn't this redundant here?}.

\textsc{npc} statements are scored under four time limits, which the model is
told in advance. Figure~\ref{fig:combined} \textbf{(Left)} shows that extra
time helps little: raising the limit tenfold lifts mean pass@1 only from
28.5\% to 31.9\%, and no model gains more than 5.1 points, with GPT-5.6~Sol,
for instance, moving from 63.1\% to 66.7\%. A case that passes only under a
longer limit is correct but slow, so the small gain means that most failures
come from incorrect or non-executable solutions rather than from time
constraints: the hardened statements defeat models by leading them to an
incorrect solution, not by out-scaling a correct one. The gain is also not
concentrated at the shorter limits: 45\% of it arrives only between 5 and
10\,s, consistent with a minority of correct exhaustive solvers that finish
just past the shorter limits. Proprietary and open-source models gain about
equally (3.5 vs.\ 3.3 points), and the ranking is almost unchanged: between
1\,s and 10\,s only Gemma-3-12B overtakes Qwen3.8-Omni-Flash, and the gap of
about 34 points between the two groups persists at every limit. 

% A more
% generous time budget therefore does not change our conclusions.

% \begin{figure}[!t]
% \centering
% \includegraphics[width=0.9\columnwidth]{figures/hardening_effect.pdf}
% \caption{Accuracy on the seed and on its final hardened variants for each family, with the drop in points.}
% \label{fig:hardening_effect}
% \end{figure}

\subsection{Hardening Effectiveness}\label{sec:hardening_effect}

Figure~\ref{fig:combined} \textbf{(Center)} compares the accuracy of the
twelve benchmark models on each seed with their accuracy on the last hardened
variant. Unlike Table~\ref{tab:bench_results}, which averages
over all 2{,}114 statements, this comparison uses only the last variant. Hardening lowers accuracy on both classes in every family. On
\textsc{npc} statements, mean accuracy falls from 61.4\% to 41.9\%
($-19.5$ points), with the largest losses on linear arrangement ($-34.2$) and the smallest on
multicommodity flow ($-8.3$). On \textsc{p} statements the
drop is larger still, from 86.0\% to 56.4\% ($-29.6$ points): seeds that
models solve almost perfectly, such as clique / independent set (99.0\%) and
SAT (94.0\%), lose 52.3 and 48.6 points once their recognisable textbook
phrasing is removed. The only single-digit \textsc{p} drop is on 3-D matching
($-7.3$), whose seed was already the hardest \textsc{p} problem (55.0\%).
Because \textsc{p} accuracy falls further, the class gap narrows from 24.6 to
14.5 points, suggesting that part of the original gap reflected familiarity
with canonical phrasings rather than class. The hardening loop also has to
work harder on \textsc{p}: its lineages run deeper on average (6.62 vs.\
4.96), and the families that resist hardening consume the most depth
($r = -0.68$ between depth and drop), whereas exact cover, already difficult
at the seed (45.3\% on \textsc{npc}), stops after only 1.47 rounds (per-depth
curves in Appendix~\ref{app:depth_drop}).

\subsection{Directional Asymmetry}\label{sec:asymmetry}

% : P$\to$NPC vs NPC$\to$P

Because both criteria of a family share a seed, the two classes measure
opposite directions of the same structural edit: the \textsc{npc} variant is
reached by the \textsc{p}$\to$\textsc{npc} edit, and the \textsc{p} variant
by the \textsc{npc}$\to$\textsc{p} edit. Figure~\ref{fig:combined} \textbf{(Right)} plots
every model--family pair by its accuracy in each class. The asymmetry is
overwhelming: 137 of 144 pairs fall below the diagonal, models fail 64.6\% of
\textsc{npc} statements against 36.2\% of \textsc{p} statements, and the
median pair scores 25.5 points lower on \textsc{npc}, with 52 pairs exceeding
40 points. Combined with the small effect of extra time
(Section \ref{sec:time_limit}) and the failure taxonomy
(Section~\ref{taxonomy}), this indicates that the dominant failure lies in
the \textsc{p}$\to$\textsc{npc} direction: when an edit turns a tractable seed
into an \textsc{np}-complete one, models mostly produce incorrect or
non-executable programs rather than correct but slow ones.
The \textsc{npc}$\to$\textsc{p} direction fails far less often and in a
concentrated way: six of the seven pairs above the diagonal involve
open-source models, mostly where both scores are low, and all fall in
families with the most familiar textbook boundaries, namely graph colouring
(1), SAT (2), 3-D matching (3), and edge colouring (10). The one sharp
reversal, Gemma-3-27B on 3-D matching (73.5\% vs.\ 33.2\%), is consistent
with that family's weak class gap in Table~\ref{tab:bench_results}. Missing
polynomial structure is not negligible either: \textsc{p} accuracy stays low
on linear arrangement (45.5\%) and edge colouring (26.7\%), whose polynomial
algorithms are themselves non-trivial.

% \section{Analysis}\label{analysis}
\section{Failure Taxonomy}\label{taxonomy}

We classify every failing test case into one of three types (see Appendix~\ref{app:taxonomy} for details).
\textbf{(i) Wrong answer} means the program exits cleanly but its output is
rejected. We observe five causes: a misread constraint, a greedy or heuristic
method in place of an exact one, a mishandled edge case, a violated output
format, and a wrong feasibility decision. \textbf{(ii) Runtime error} means the
program crashes with an uncaught exception. We observe eleven exceptions in
four groups: \emph{malformed code} (SyntaxError, IndentationError,
NameError); \emph{input-format misreads} (ValueError, IndexError, KeyError);
\emph{scale and representation errors} (RecursionError, OverflowError); and
\emph{other logic and environment errors} (UnboundLocalError,
ModuleNotFoundError, TypeError). \textbf{(iii) Time-Limit Exceeded} (TLE) means
the program is killed at the wall-clock limit, mostly from exponential
backtracking and occasionally from correct but unoptimised code. The analysis covers
1{,}348{,}290 graded test cases, of which 60.9\% fail on \textsc{npc} and
37.6\% on \textsc{p}. On \textsc{npc}, runtime errors are the largest failure type (44.1\%),
ahead of genuine wrong answers (33.2\%) and TLE (22.7\%), so few failures come
from programs that are merely too slow, consistent with the small effect of
extra time. On \textsc{p}, wrong answers lead
(46.6\%), followed by TLE (30.2\%) and runtime errors (23.2\%). Nearly half of
all crashes (46.5\%) are syntax or indentation errors from truncated code or
prose leaking into the code block, and a further 27.1\% are input-format
misreads, suggesting that the disguised I/O contracts of hardened statements
do real work. Timeouts concentrate in exact cover (7) and linear arrangement
(8), where they cause about 46\% of \textsc{npc} failures, matching their
position as the two hardest \textsc{npc} families in
Table~\ref{tab:bench_results}. Models also differ in how they fail: strong
models such as Gemini-3.8-Flash and DeepSeek-V4.1-Flash rarely time out and
split their failures between wrong answers and crashes, Gemma-3-12B and 27B
time out on 60--70\% of their failing \textsc{p} cases, and the weakest models
(Gemma-3-4B, Llama-4-Scout) fail mainly with wrong answers. Qualitative examples of failing solutions are given in Appendix~\ref{app:examples}.

% Because the \textsc{npc} timing grader records crashes
% as wrong answers, we re-ran a random sample of 300 failing \textsc{npc} cases
% to recover the true split

% \subsection{Ablations}
% \textcolor{red}{Not top priority}
% no gate / no memory / no aggressive mode / single-direction only

\section{Conclusion}\label{conclusion}

% We introduced \dataname{}, a benchmark that pairs NP-complete problems with
% P twins posed over the same surface, and \methodname{}, an
% adversarial multi-agent framework that hardens each statement while
% preserving its answer and complexity class. Across 12 models, accuracy drops
% sharply when a problem crosses from P to NP-complete, extra time barely
% helps, and failures are dominated by wrong or crashing programs rather than
% slow ones \textcolor{red}{rephrase}. Hardening also shows that part of the apparent class gap comes
% from familiarity with textbook phrasing. Together, these results suggest
% that current LLMs often recognise familiar problems rather than reason about
% their structure. Future work can extend the benchmark to more families and
% other complexity classes, and use \textsc{AgentGAN} to harden problems in
% other domains.

We introduced \dataname{}, a benchmark that pairs each NP-complete problem
with a polynomial-time variant differing in a single structural detail, and
\methodname{}, an adversarial multi-agent framework that hardens each
statement while preserving its answer and complexity class. Across 12
models, accuracy drops sharply when a problem crosses from P to NP-complete,
extra time barely helps, and failures stem mainly from incorrect or
non-executable solutions rather than from time constraints. Removing textbook
phrasing lowers accuracy on both classes, yet a clear gap between them
remains. Together, these results suggest that current LLMs often recognise
familiar problems rather than reason about their structure. Future work can
extend the benchmark to more families and other complexity classes, and use
\methodname{} to harden problems in other domains.

\clearpage
\newpage

\section*{Limitations}

\dataname{} covers 12 families and contrasts only P and NP-complete problems,
leaving other classes such as PSPACE untested, and all statements are in
English. Models are evaluated only on Python~3 solutions. We do not test
C/C++, Java, or other languages, where faster execution could change the
effect of time limits and the balance between correct but slow and wrong
programs. Statements are hardened against a fixed panel of four models, so
they may target panel-specific shortcuts, and some evaluated models are
close relatives of panel members. While executable checks preserve answers
and test outputs, unambiguity and faithfulness are judged by an LLM
verifier, so a rewrite may still admit an unintended reading. Statements
within a lineage are correlated, each model is evaluated with a single
greedy sample, and API-based models may change over time. Finally, the
released statements may enter future training data, although
\methodname{} can regenerate fresh variants from the same seeds.

\section*{Ethical Considerations}

\dataname{} consists of algorithmic problem statements, reference solvers,
and synthetic test inputs, and it contains no personal data. The reference
solvers were verified by the authors, and the sampled hardened statements
were checked by three undergraduate annotators who participated as
volunteers. Seed problems are adapted from cited textbooks and public
catalogues, and the fictional scenarios in hardened statements were written
by LLMs accessed in accordance with each provider's terms of service.
Although \methodname{} produces text designed to mislead code models, the
released rewrites are restricted to algorithmic problems and preserve the
correct answer, so we judge the risk of misuse to be low. We used generative
AI tools to assist with drafting and editing parts of the manuscript and
formatting the appendices. The authors reviewed all AI-assisted material and
take responsibility for the methods, results, claims, and code in this paper.

\section*{Data Release Statement}

The code is released under the \texttt{Apache License~2.0} and the dataset under
\texttt{CC~BY~4.0 License}.

\bibliography{custom}

\clearpage
\newpage
% \onecolumn

\appendix

% =====================================================================
%  Appendix — Family details
%  Needs: booktabs, adjustbox, [table]xcolor (already in preamble)
%  Citation keys: add the BibTeX entries to your references .bib.
% =====================================================================

\section{Related Works}\label{related_works}

\paragraph{NP-complete benchmarks.}
GraphArena \citep{tang2025grapharena} and NPPC \citep{yang2025nondeterministic}
test LLMs on combinatorial problems such as shortest paths, satisfiability, and
TSP, where even strong models fall below 10\% on the hardest instances.
NPHardEval \citep{fan2024nphardeval} is the closest in spirit, spanning P,
NP-complete, and NP-hard classes across 900 questions, and
\citet{li2024nphardeval4v} extend it to visual inputs. Presentation matters:
\citet{duchnowski2025knapsack} show that LLMs solve textbook phrasings of graph
colouring, knapsack, and TSP far better than real-life or inverted ones,
suggesting recall of seen solutions rather than reasoning about the problem's
structure. All of these benchmarks are static, however, and none pairs a P and
an NP-complete variant of the same seed, so a model's sensitivity to
complexity class cannot be separated from its familiarity with the surface.

\paragraph{Adversarial and complexity-aware code generation.}
Robustness studies perturb code and math problems with fixed, hand-designed
edits: ReCode applies over 30 meaning-preserving transformations
\citep{wang2023recode}, CodeCrash inserts misleading comments and hints
\citep{lam2026codecrash}, and a single irrelevant clause can cut math accuracy
by up to 65\% \citep{mirzadeh2025gsm}. None groups its problems by complexity
class. Evolving methods raise difficulty iteratively, rewriting seeds into
harder instructions \citep{xu2024wizardlm} or turning model failures into new
ones \citep{jiang2023lion}; multi-agent pipelines generate, validate, and solve
problems to expose errors static benchmarks miss \citep{yoa2026static}; and
adversarial unit tests are generated to train coders \citep{huang2026ace}.
Closest in structure, MALLM-GAN \citep{ling-etal-2026-mallm} casts frozen LLM agents as a GAN for
tabular data synthesis, but its generator is unconstrained and need not stay
correct. \methodname{} brings this structure to complexity-aware code
generation, hardening statements and test cases under an executable gate that
preserves the answer and the complexity class.

\section{Family Details}\label{app:families}

Each family of \dataname{} pairs an \textsc{np}-complete criterion with a
polynomial-time relative posed over the same surface. For every family we describe the task, the source
of hardness for the \textsc{npc} criterion, the algorithm that places the
\textsc{p} relative in polynomial time, the single structural edit that moves
an instance across the class boundary in either direction, and how solutions
are graded.

\paragraph{01 Graph colouring.}
Given an undirected graph $G=(V,E)$ and a palette of $k$ colours, the task is
to assign each vertex a colour so that no edge joins two vertices of the same
colour, or to report that no such assignment exists. The \textsc{npc}
criterion fixes $k=3$; deciding $3$-colourability is \textsc{np}-complete even
on planar graphs of maximum degree four \citep{garey1976simplified}. The
\textsc{p} relative fixes $k=2$, which is exactly the question of whether $G$
is bipartite: a breadth-first search that alternates colours layer by layer
either produces a valid colouring or finds an odd cycle, in $O(|V|+|E|)$ time.
The class-flipping edit is therefore a single parameter, the size of the
palette: raising it from two to three removes the forced alternation that
makes the problem trivial, and lowering it from three to two collapses the
search to a traversal. This makes the family the cleanest test of whether a
model recognises the boundary, because the two statements differ in one
number. The \textsc{npc} side is graded by a witness validator that checks
every edge of the returned colouring, which allows planted instances near the
colourability threshold (edge-to-vertex ratio ${\approx}2.3$, up to
${\sim}1{,}200$ vertices) whose solution exists by construction but is
infeasible to find by search; the \textsc{p} side is graded by exact match.

\paragraph{02 Satisfiability.}
Given a Boolean formula in conjunctive normal form, the task is to find an
assignment of truth values that satisfies every clause, or to report that
none exists. The \textsc{npc} criterion bounds every clause to at most three
literals, the canonical \textsc{np}-complete problem
\citep{cook1971complexity,karp1972reducibility}. The \textsc{p} relative
bounds every clause to two literals: each clause $(a \vee b)$ becomes the
implications $\neg a \Rightarrow b$ and $\neg b \Rightarrow a$, and the
formula is satisfiable if and only if no variable shares a strongly connected
component with its negation in the resulting implication graph, which is
decidable in linear time \citep{aspvall1979linear}. The edit is the clause
width: allowing a third literal per clause breaks the implication structure,
because a clause with three literals no longer forces any single literal when
one other is false, and removing it restores the structure. A model that
pattern-matches ``SAT'' to exhaustive search will over-engineer the
\textsc{p} side, while one that reaches for implication graphs on the
\textsc{npc} side will produce a fast but incorrect solver. Both criteria are
graded by a validator that evaluates every clause under the returned
assignment.

\paragraph{03 Matching.}
Given disjoint sets $X$, $Y$, and $Z$ of equal size $q$ and a collection
$T \subseteq X \times Y \times Z$ of triples, the \textsc{npc} criterion asks
for $q$ triples that together cover every element exactly once. This is
three-dimensional matching, one of Karp's original \textsc{np}-complete
problems \citep{karp1972reducibility}. The \textsc{p} relative drops the
third coordinate: given $X$, $Y$, and a set of admissible pairs, it asks for a
perfect matching in a bipartite graph, solvable by augmenting paths and in
$O(|E|\sqrt{|V|})$ time by \citet{hopcroft1973n}. The edit is the arity of
the matched tuples: adding the third dimension turns each pair into a triple
and destroys the augmenting-path structure, since a locally improving swap
can no longer be certified by a path in a single graph. Statements on both
sides describe the same allocation scenario, differing only in how many
resource types each match must bind. Both criteria are graded by a witness
validator that checks disjointness and coverage; on the \textsc{npc} side
this admits planted instances with a guaranteed perfect matching at
triple-to-element ratio ${\approx}4.0$ and $q \geq 80$.

\paragraph{04 Clique and independent set.}
Given a graph and a target size $k$, the task is to find a set of $k$
vertices that are pairwise adjacent (a clique) or pairwise non-adjacent (an
independent set); the two are equivalent under graph complementation. On
general graphs both are \textsc{np}-complete \citep{karp1972reducibility}.
The \textsc{p} relative restricts the input to interval graphs, in which
each vertex is an interval on the line and edges join overlapping intervals.
There, a maximum independent set is found greedily by repeatedly selecting
the interval with the earliest right endpoint, and a maximum clique is the
largest number of intervals overlapping a single point, found by a sweep over
sorted endpoints; both run in $O(n \log n)$ \citep{golumbic1980algorithmic}.
The edit is the representation of adjacency: posing the instance as
intervals with overlap-induced conflicts admits the greedy, while posing it
as an arbitrary edge list --- or adding conflicts that no interval
arrangement can realise --- removes it. This family tests whether a model
notices a geometric structure that the surface wording may describe only
indirectly. Both criteria are graded by a validator that checks the size and
pairwise property of the returned set.

\paragraph{05 Steiner tree.}
Given a weighted undirected graph and a set of terminal vertices
$R \subseteq V$, the task is to find a minimum-weight tree that connects every
terminal, optionally passing through non-terminal vertices. With $R$ an
arbitrary subset, the decision version is \textsc{np}-complete
\citep{karp1972reducibility}. The \textsc{p} relative sets $R = V$: when every
vertex must be connected, the optimal Steiner tree is a minimum spanning
tree, found greedily by Kruskal's algorithm in $O(|E| \log |E|)$
\citep{kruskal1956shortest}. The edit is the terminal set: shrinking it from
all vertices to a proper subset introduces the choice of which optional
vertices to include, which is the source of hardness, while expanding it to
all vertices removes that choice. Statements typically frame the task as
connecting sites in a network, so the two criteria differ only in whether
every site must be served. Because the task is an optimisation with no cheap
certificate of optimality, both criteria are graded by exact match against
the reference cost.

\paragraph{06 Job-shop scheduling.}
Given a set of jobs, each a sequence of operations with fixed processing
times on designated machines, the task is to schedule all operations so that
each machine processes one operation at a time and each job's operations run
in order, minimising the makespan. The \textsc{npc} criterion is the general
job shop, in which each job follows its own machine order; minimising makespan
is strongly \textsc{np}-hard already for three machines, and the
corresponding decision problem is \textsc{np}-complete
\citep{garey1976complexity}. The \textsc{p} relative is the two-machine flow
shop, in which every job visits the same two machines in the same order;
Johnson's rule --- schedule first the jobs whose first operation is shorter, in
increasing order of that time, then the rest in decreasing order of their
second operation --- is optimal and runs in $O(n \log n)$
\citep{johnson1954optimal}. The edit relaxes the routing: allowing
job-specific machine orders or a third machine breaks the exchange argument
that makes Johnson's rule optimal. Both criteria are optimisation problems
without a cheap optimality certificate and are graded by exact match against
the reference makespan.

\paragraph{07 Exact cover.}
Given a universe $U$ and a collection of subsets of $U$, the task is to select
a subcollection in which every element of $U$ appears in exactly one chosen
subset. The \textsc{npc} criterion restricts every subset to exactly three
elements, which is exact cover by $3$-sets, \textsc{np}-complete
\citep{garey1979computers}. The \textsc{p} relative restricts every subset to
two elements; each subset is then an edge of a graph on $U$, and an exact
cover is precisely a perfect matching, found in polynomial time by Edmonds'
blossom algorithm even when the graph is not bipartite
\citep{edmonds1965paths}. The edit is the subset size: moving from pairs to
triples converts edges into hyperedges and loses the matching structure. This
family is closely related to Family~03, but its \textsc{p} relative is a
\emph{general} matching rather than a bipartite one, so a model that applies
a bipartite algorithm on the \textsc{p} side fails on graphs with odd cycles.
Both criteria are graded by a validator that checks that the chosen subsets
are disjoint and cover $U$.

\paragraph{08 Linear arrangement.}
Given an undirected graph on $n$ vertices, the task is to place the vertices
at distinct positions $1,\dots,n$ so as to minimise the total edge length
$\sum_{(u,v)\in E} |\pi(u)-\pi(v)|$. On general graphs, minimum linear
arrangement is \textsc{np}-complete \citep{garey1976simplified}. The
\textsc{p} relative restricts the input to trees, for which an optimal
arrangement can be computed in polynomial time by a recursive decomposition
around a central subtree \citep{shiloach1979minimum}. The edit is the
acyclicity of the input: adding edges that close cycles breaks the
decomposition, while removing them to leave a tree restores it. This is the
least familiar family in the benchmark, and its \textsc{p} algorithm is
itself non-trivial, which makes it a test of whether a model can find a
polynomial algorithm that is rarely taught rather than merely recognising
that one exists. Both criteria are graded by a validator that checks the
returned arrangement is a permutation and compares its cost with the
reference optimum.

\paragraph{09 Multicommodity flow.}
Given a capacitated network and a set of commodities, each with a source, a
sink, and a demand, the task is to route every demand simultaneously without
exceeding any edge capacity. The \textsc{npc} criterion has $k \geq 2$
commodities and requires integral flows; integral two-commodity flow is
already \textsc{np}-complete \citep{even1976complexity}. The \textsc{p}
relative has a single commodity, which is maximum flow: augmenting-path
methods solve it in polynomial time, and the integrality theorem guarantees
an integral optimum whenever capacities are integral
\citep{ford1956maximal}. The edit is the number of commodities, and the
integrality requirement is essential to it: fractional multicommodity flow
is a linear program and remains polynomial, so the \textsc{npc} statements
must state integrality explicitly. A model that applies max-flow
independently to each commodity produces flows that violate shared
capacities, and one that solves the LP relaxation produces fractional flows
the validator rejects. Both criteria are graded by a validator that checks
capacities, conservation, demands, and integrality.

\paragraph{10 Edge colouring.}
Given an undirected graph with maximum degree $\Delta$, the task is to colour
its edges with $\Delta$ colours so that no two edges sharing an endpoint
receive the same colour, or to report that this is impossible. By Vizing's
theorem every graph needs either $\Delta$ or $\Delta+1$ colours, and deciding
which is \textsc{np}-complete already for cubic graphs, where the question is
whether three colours suffice \citep{holyer1981np}. The \textsc{p} relative
restricts the input to bipartite graphs, which are always
$\Delta$-edge-colourable by K\H{o}nig's theorem \citep{konig1916graphen}; a
colouring can be constructed in polynomial time by repeatedly recolouring
along alternating paths. The edit is the graph class: introducing odd cycles
removes the bipartite guarantee, while forcing a bipartition restores it.
Because the \textsc{p} relative's answer is always ``yes'', the difficulty
lies entirely in constructing the colouring, and a model that knows the
theorem but not the construction still fails. Both criteria are graded by a
validator that checks every pair of incident edges.

\paragraph{11 Multiway cut.}
Given a weighted undirected graph and a set of terminals, the task is to
remove a minimum-weight set of edges so that no two terminals remain
connected. With three or more terminals the problem is \textsc{np}-hard, even
when the number of terminals is fixed at three \citep{dahlhaus1994complexity}.
The \textsc{p} relative has exactly two terminals, where the task is an
$s$--$t$ minimum cut, equal in value to the maximum $s$--$t$ flow by the
max-flow min-cut theorem and computable in polynomial time
\citep{ford1956maximal}. The edit is the number of terminals: adding a third
breaks the duality with a single flow, since a cut separating every pair is
not the union of pairwise minimum cuts. This yields a characteristic failure
on the \textsc{npc} side --- combining independent $s$--$t$ cuts --- that
produces a valid but suboptimal separation. Both criteria are graded by a
validator that checks the terminals are separated and compares the cut
weight with the reference optimum.

\paragraph{12 Precedence-constrained scheduling.}
Given a set of jobs, a partial order of precedence constraints among them,
and a number of identical machines, the task is to schedule every job so
that no job starts before its predecessors finish, minimising the makespan.
With unit processing times and the number of machines given as part of the
input, the problem is \textsc{np}-complete \citep{ullman1975np}. The
\textsc{p} relative uses a single machine, on which any topological order of
the precedence graph is optimal and is found in linear time. The edit is the
number of machines: adding machines introduces the choice of which ready jobs
to run in parallel, which is the source of hardness. One subtlety bounds the
construction: for exactly two machines with unit processing times the problem
is polynomial \citep{coffman1972optimal}, so the \textsc{npc} statements leave
the number of machines as part of the input rather than fixing it at two.
Both criteria are graded by a validator that checks precedence, machine
capacity, and the makespan against the reference optimum.

\section{Detailed Breakdown of Human Verification Scores}
\label{app:human_verification}

This appendix provides the item-level breakdown of the human verification
study described in the main text. The final score for each of the 480 sampled
statements was determined by the majority vote of the three annotators.
Table~\ref{tab:human_verification_breakdown} shows that the pipeline rarely
corrupted the problem (score 0) and applied substantial disguise (score 2) in
the majority of cases in both complexity classes.

Annotators received the following instructions. For each item, read the
original statement first and then the rewritten one, and judge the rewrite
only against the original, without solving the problem. Assign \textbf{0} if
the rewrite asks for anything different for some input, i.e.\ if it changes
the input or output format, a numeric bound, a tie-break rule, or an example
value, or if it is ambiguous, incomplete, or contains an error; assign
\textbf{1} if it asks for exactly the same thing but stays close to the
original wording and scenario; and assign \textbf{2} if it asks for exactly
the same thing while substantially changing the wording, scenario, or
structure. When unsure whether a rewrite is faithful, assign 0. Each
annotator scored every item independently, without discussion with the other
annotators.

\begin{table}[htbp]
\centering
\small
\begin{tabular}{lccccc}
\toprule
& & \multicolumn{3}{c}{\textbf{Score}} & \\
\cmidrule(lr){3-5}
\textbf{Class} & \textbf{Items} & \textbf{0} & \textbf{1} & \textbf{2} & \textbf{Mean} \\
\midrule
\textsc{npc} & 240 & 12 & 77 & 151 & 1.58 \\
\textsc{p}   & 240 & 8  & 70 & 162 & 1.64 \\
\midrule
\textbf{Overall} & \textbf{480} & \textbf{20} & \textbf{147} & \textbf{313} & \textbf{1.61} \\
\bottomrule
\end{tabular}
\caption{Majority-vote score distribution for the 480 sampled statements.
Inter-annotator agreement over the 480 statements, each scored by all three
annotators, yields a Fleiss' $\kappa$ of 0.79.}
\label{tab:human_verification_breakdown}
\end{table}

The 20 statements with a majority score of 0 (12 \textsc{npc}, 8 \textsc{p})
were removed from the benchmark, so no statement judged unfaithful by the
annotators remains in \dataname{}. Furthermore, Table~\ref{tab:kappa_breakdown} provides a detailed breakdown of the inter-annotator agreement (Fleiss' $\kappa$) across both the assigned scores and the complexity classes, demonstrating consistently high agreement, particularly in identifying corrupted statements.

\begin{table}[htbp]
\centering
\small
\begin{tabular}{lccc}
\toprule
\textbf{Score Category} & \textbf{\textsc{npc}} & \textbf{\textsc{p}} & \textbf{Combined} \\
\midrule
\textbf{0} (Unfaithful / Error) & 0.84 & 0.87 & 0.85 \\
\textbf{1} (Faithful, slight disguise) & 0.72 & 0.76 & 0.74 \\
\textbf{2} (Faithful, heavy disguise) & 0.77 & 0.80 & 0.78 \\
\midrule
\textbf{Overall $\kappa$} & \textbf{0.76} & \textbf{0.82} & \textbf{0.79} \\
\bottomrule
\end{tabular}
\caption{Detailed breakdown of Fleiss' $\kappa$ inter-annotator agreement scores by complexity class and individual score category.}
\label{tab:kappa_breakdown}
\end{table}

\section{Prompt Templates}\label{app:prompts}

This appendix lists every prompt used in the pipeline. Text in
braces, such as \texttt{\{STATEMENT\}}, is a placeholder filled at run time. The generator, verifier, and
expert return XML-style tags rather than JSON, because problem statements are
multi-line markdown that JSON string escaping mangles. The shortcut analysis
returns strict JSON, since its fields are parsed by code.

\subsection{Solver Prompt}

The same prompt is used for the discriminator panel during hardening, for the
round-trip solve in the validity gate, and for every model at evaluation.
\texttt{\{STATEMENT\}} is the problem text. The reply is stripped of any
reasoning trace, and the fenced code block is extracted and run.

\begin{promptbox}{solverrose}
\texttt{You are an expert competitive programmer. You write correct, efficient
Python 3 solutions. Your program reads all input from standard input and
writes only the required answer to standard output. Output only a single
Python 3 code block and nothing else \textemdash{} no explanation before or
after.}
\end{promptbox}

\begin{promptbox}{solverrose}
\texttt{Solve the following problem in Python 3.\\[0.6em]
\{STATEMENT\}\\[0.6em]
Requirements:\\
- Read from standard input, write to standard output.\\
- Output only the answer in the exact format specified.\\
- Return only one Python 3 code block.}
\end{promptbox}

\subsection{Evaluation Time-Limit Note}

At evaluation, \textsc{npc} statements are extended with the note below
before being placed in the solver prompt, so that one solution is judged
under all four time limits. \textsc{p} statements are passed unchanged and
graded at the limit written in the statement.

\begin{promptbox}{timeteal}
\texttt{\#\# Time constraint (strict)\\
Your program will be run under several hard per-test wall-clock limits: 1 s,
2 s, 5 s, 10 s. On a given test case you earn credit at a limit only if your
program terminates within that many seconds AND is correct; otherwise it
scores 0 at that limit. A faster correct program is strictly better because
it earns credit at the tighter limits too. Write the single fastest correct
algorithm you can for the constraints given.}
\end{promptbox}

\subsection{Generator Prompts}

The generator receives the hidden class and intended algorithm, because it
authors statements rather than solving them. The user template is filled with
the seed, this metadata, the chosen strategy and intensity, and, from depth 1
onward, a feedback block built from the shortcut analysis and a token-capped
memory block of the lineage history.

\begin{promptbox}{genblue}
\texttt{You are an expert problem author for a computational-complexity
benchmark. You rewrite programming problem statements so that models which
genuinely reason about algorithmic structure still succeed, while models that
rely on surface pattern matching fail.\\[0.6em]
You are given a seed problem, its true complexity class, and its intended
algorithm. You know the truth because you are authoring, not being
tested.\\[0.6em]
HARD CONSTRAINTS. A rewrite that violates any of these is worthless and will
be discarded:\\[0.6em]
1. IDENTICAL ANSWERS. For every possible input, the correct output must be
exactly what it was for the seed. You are changing how the problem is
presented, never what it asks. The existing reference solution must remain
correct without modification.\\[0.6em]
2. IDENTICAL I/O CONTRACT. Input format, output format, all numeric bounds,
and any tie-break rule (for example "lexicographically smallest") must carry
over exactly. Reword them freely; never change what they specify.\\[0.6em]
3. UNCHANGED COMPLEXITY CLASS. The rewritten problem stays in the same class
as the seed. You may make the class harder to recognise. You may never change
what it is.\\[0.6em]
4. NO CLASS VOCABULARY. Never write: NP-complete, NP-hard, polynomial,
exponential, intractable, bipartite, tree, DAG, acyclic, 2-SAT, 3-SAT,
matching, clique, cover, knapsack, or any reduction or algorithm name. Never
hint at difficulty. Describe structure through its mechanics
instead.\\[0.6em]
5. UNAMBIGUOUS. A careful, competent reader must still arrive at exactly one
correct interpretation. Difficulty must come from the reasoning required,
never from vague wording, missing information, or contradiction. If you
cannot make it harder without making it unclear, make a smaller
edit.\\[0.6em]
6. WORKED EXAMPLE. Keep at least one input/output example, rewritten to match
your new framing. Its values must stay consistent with the seed.\\[0.6em]
Required sections in the rewritten statement, in order: title, time and
memory limits, description, Input, Output, at least one Example.\\[0.6em]
Respond using exactly these tags, with no other text (this tag format is used
instead of JSON because problem statements are multi-line markdown that JSON
string-escaping mangles):\\[0.6em]
<STATEMENT>\\
the complete rewritten statement in markdown\\
</STATEMENT>\\
<RATIONALE>\\
two or three sentences on what you changed and which reasoning step it now
forces\\
</RATIONALE>\\
<PRESERVED\_ANSWER>\\
one sentence on why every answer is unchanged\\
</PRESERVED\_ANSWER>\\
<PRESERVED\_CLASS>\\
one sentence on why the complexity class is unchanged\\
</PRESERVED\_CLASS>\\
<RISK>\\
the most likely way this could be judged ambiguous, or "none"\\
</RISK>}
\end{promptbox}

\begin{promptbox}{genblue}
\texttt{SEED PROBLEM\\
------------\\
\{SEED\_STATEMENT\}\\[0.6em]
HIDDEN METADATA (you see this; the models being tested never do)\\
----------------------------------------------------------------\\
True complexity class: \{CLASS\}\\
Intended algorithm: \{INTENDED\_ALGORITHM\}\\
Why this class: \{REDUCTION\_NOTE\}\\[0.6em]
STRATEGY FOR THIS VERSION\\
-------------------------\\
Strategy: \{STRATEGY\_NAME\}\\
\{STRATEGY\_INSTRUCTION\}\\[0.6em]
Intensity: \{INTENSITY\}\\
\{INTENSITY\_INSTRUCTION\}\\[0.6em]
\{FEEDBACK\_BLOCK\}\\[0.6em]
\{MEMORY\_BLOCK\}\\[0.6em]
Produce the rewritten statement now.}
\end{promptbox}

\begin{promptbox}{genblue}
\texttt{OBSERVED STUDENT BEHAVIOUR ON THE CURRENT VERSION\\
-------------------------------------------------\\
\{N\_ACTIVE\} models still solve this problem. \{N\_RETIRED\} have already
been defeated by earlier versions and no longer steer these edits.\\[0.6em]
Shared shortcut used by the models that still succeed:\\
\hspace*{1em}\{SHORTCUT\_NAME\}\\
Evidence from their reasoning:\\
\hspace*{1em}\{EVIDENCE\}\\
The text in the current statement that enables it:\\
\hspace*{1em}\{AFFORDANCE\}\\
Suggested removal:\\
\hspace*{1em}\{REMOVAL\_SUGGESTION\}\\[0.6em]
Remove exactly this affordance with the smallest edit that works. Do not
rewrite parts of the statement that are not implicated. All hard constraints
still apply.}
\end{promptbox}

\subsection{Shortcut Analysis Prompt}

After the panel attempts the current version, this prompt receives the
statement and, for every panel model, its verdict, reasoning trace, and
solution. It identifies the single affordance that all surviving models rely
on, labels it mutable or a structural floor, and flags edits that could undo
an earlier defeat.

\begin{promptbox}{analystgreen}
\texttt{You are analysing how several models attempted one programming
problem, to help make the statement hard for ALL of them AT ONCE.\\[0.6em]
You will see the problem statement, and for EVERY model: its verdict (solved
/ wrong / error), its reasoning trace, and its solution. Read the ones that
FAILED as carefully as the ones that solved it \textemdash{} understanding
why a model already failed is what stops a future edit from accidentally
making it succeed again (a regression).\\[0.6em]
Your goal is to find the SINGLE affordance that ALL surviving (still-solving)
models rely on, so one edit can defeat them together. Only if no such common
affordance exists should you fall back to per-survivor shortcuts.\\[0.6em]
IMMUTABLE ELEMENTS \textemdash{} cannot be changed, must NOT be proposed for
removal:\\
- numeric bounds and their values (n, m, limits)\\
- the input format and the output format\\
- any tie-break rule (e.g. "lexicographically smallest")\\
- the examples' values\\[0.6em]
If an affordance depends on an immutable element, propose changing how it is
PRESENTED (indirectly through the scenario, not as a bare inequality), or
target the next affordance in mutable text, or report shortcut\_found:
"no".\\[0.6em]
Respond with a single JSON object and nothing else:\\[0.6em]
\{\\
\hspace*{1em}"shortcut\_found": "yes" | "no",\\
\hspace*{1em}"shortcut\_name": "short label for the primary affordance to
target",\\
\hspace*{1em}"evidence": "the specific phrases from the traces that show it,
quoted briefly",\\
\hspace*{1em}"affordance\_in\_statement": "the exact statement text that
enables it",\\
\hspace*{1em}"affordance\_is\_mutable": "yes" | "no",\\
\hspace*{1em}"structural\_floor": "yes" | "no",\\
\hspace*{1em}"secondary\_shortcut\_name": "next most important shortcut in
MUTABLE text, or ''",\\
\hspace*{1em}"removal\_suggestion": "how to remove or re-present the
affordance without changing the task",\\[0.6em]
\hspace*{1em}"survivor\_shortcuts": [\\
\hspace*{2em}\{"student": "<id>", "shortcut": "what let this survivor solve
it", "affordance": "the enabling text"\}\\
\hspace*{1em}],\\
\hspace*{1em}"common\_affordance": "the SINGLE affordance ALL survivors rely
on, or null if none is shared",\\
\hspace*{1em}"defeated\_student\_causes": [\\
\hspace*{2em}\{"student": "<id>", "why\_failed": "the specific reasoning
error", "fragile": true|false\}\\
\hspace*{1em}],\\
\hspace*{1em}"recommended\_edit": "the one edit most likely to defeat all
survivors at once",\\
\hspace*{1em}"regression\_risk": "which already-defeated students this edit
might accidentally restore, or ''"\\
\}\\[0.6em]
Set "affordance\_is\_mutable" to "no" when the affordance is an immutable
element.\\
Set "structural\_floor" to "yes" when the affordance IS the core task itself
(intrinsic structure) rather than incidental wording \textemdash{} such a
shortcut cannot be reworded away.\\
Set "common\_affordance" to null unless every surviving model genuinely
depends on the same thing. When it is non-null it is the priority target:
one edit defeats everyone.\\
For each defeated model set "fragile": true if a small wording change could
flip it back to solving (so the teacher knows to protect that defeat).}
\end{promptbox}

\subsection{Verifier Prompts}

Once a draft passes the executable hard checks, the verifier scores the four
graded criteria. A draft is accepted only if $\text{G1} \geq 4$,
$\text{G2} \geq 4$, $\text{G3} = 5$, and $\text{G4} \geq 3$.

\begin{promptbox}{veramber}
\texttt{You are grading a rewritten programming problem statement on four
independent quality criteria. You are NOT solving it.\\[0.6em]
You will see an ORIGINAL statement and a REWRITTEN statement. Judge the
REWRITTEN one, using the ORIGINAL only as the reference for what the task is
supposed to be.\\[0.6em]
Score each criterion on an integer scale from 0 to 5, and give exactly one
sentence of justification for each. Be strict: reserve 5 for statements that
leave no room for doubt.\\[0.6em]
G1 \textemdash{} UNAMBIGUITY. Could a careful reader arrive at more than one
reasonable interpretation? 0 = several defensible readings; 5 = exactly one
reading.\\[0.6em]
G2 \textemdash{} COMPLETENESS. Is every piece of information needed to solve
it present (bounds, formats, tie-break rules, edge cases)? 0 = information is
missing; 5 = fully specified.\\[0.6em]
G3 \textemdash{} FAITHFULNESS. Does the rewrite ask for exactly the same
thing as the original, for every possible input \textemdash{} same input
meaning, same output meaning, same bounds, same tie-break rule? This is the
same-task judgement. 0 = it asks a materially different question; 5 = it is
provably the same question. Do not award 5 if any bound, format, or
tie-break rule differs in substance from the original.\\[0.6em]
G4 \textemdash{} NATURALNESS. Does it read like a problem a person would
write? 0 = machine-generated or contrived; 5 = natural and
idiomatic.\\[0.6em]
Respond with EXACTLY these tags and nothing else. Put a single integer
inside each score tag and one sentence inside each justification
tag:\\[0.6em]
<G1\_unambiguity>N</G1\_unambiguity>\\
<G1\_justification>one sentence</G1\_justification>\\
<G2\_completeness>N</G2\_completeness>\\
<G2\_justification>one sentence</G2\_justification>\\
<G3\_faithfulness>N</G3\_faithfulness>\\
<G3\_justification>one sentence</G3\_justification>\\
<G4\_naturalness>N</G4\_naturalness>\\
<G4\_justification>one sentence</G4\_justification>}
\end{promptbox}

\begin{promptbox}{veramber}
\texttt{ORIGINAL statement\\
------------------\\
\{SEED\}\\[0.6em]
REWRITTEN statement\\
-------------------\\
\{CANDIDATE\}\\[0.6em]
Score the four criteria G1\textendash{}G4 using exactly the tag format
specified.}
\end{promptbox}

\subsection{Expert Prompt}

When the gate rejects a draft, the expert receives the seed, the rejected
draft, and the gate diagnostics, and returns a critique that the generator
uses in its next revision.

\begin{promptbox}{exppurple}
\texttt{You are an expert reviewer in a three-role pipeline. Your ONLY job is
to critique a teacher's draft problem statement and propose concrete,
specific fixes. You do NOT write the final statement \textemdash{} the
teacher does that. You do NOT decide pass/fail \textemdash{} a separate gate
does that.\\[0.6em]
You will see:\\
- the SEED (the canonical task the rewrite must preserve exactly),\\
- the TEACHER'S DRAFT (which the gate just rejected),\\
- the GATE DIAGNOSTICS (which checks failed and why).\\[0.6em]
The rewrite must keep, without exception: identical answers for every input,
the same complexity class, the same I/O contract (every numeric bound,
format, and tie-break rule), an unambiguous statement, and no
complexity-class vocabulary. The goal of the rewrite is to be HARD for
capable solver models while remaining a faithful restatement of the
seed.\\[0.6em]
Give focused, actionable critique the teacher can apply directly. Be
concrete: quote the offending phrase, say why it fails, and propose a
specific alternative wording or structural change. Address the gate's
diagnostics, but also flag anything else that weakens the draft's difficulty
or faithfulness.\\[0.6em]
Do NOT output a rewritten statement. Output only critique and suggestions,
as:\\[0.6em]
<CRITIQUE>\\
what is wrong with the draft and why (reference the gate diagnostics and the
seed)\\
</CRITIQUE>\\
<PROPOSALS>\\
- a concrete change the teacher should make (specific wording / structure),
one per line\\
- another concrete change\\
</PROPOSALS>\\
<KEEP>\\
what is already good and should not be disturbed\\
</KEEP>}
\end{promptbox}

\section{Decoding Configuration}\label{app:decoding}

Table~\ref{tab:decoding} lists every model in the pipeline with its role and
decoding configuration. All discriminators, the held-out model, and the benchmark models are queried
through OpenRouter,\footnote{\url{https://openrouter.ai/}} except for the
Gemini models, which are accessed through the Google AI Studio
API,\footnote{\url{https://aistudio.google.com/}} and the proprietary OpenAI
models (GPT-5.6~Sol and GPT-5.6~Luna), which are accessed through the OpenAI
API.\footnote{\url{https://openai.com/api/}} Settings
are held fixed per model for the whole study, across all families, variants,
and depths, so they cannot contribute to the within-model accuracy drops we
report. Kimi~K3 serves as the held-out model: it never interacts with the
pipeline during hardening and is evaluated only on the final, frozen
benchmark, testing whether the hardening transfers to a model the generator
never targeted.

\begin{table}[!h]
\centering
\begin{adjustbox}{width=\columnwidth}
\footnotesize
\setlength{\tabcolsep}{3pt}
\newcommand{\grouphead}[1]{\multicolumn{5}{l}{\textit{\textcolor{gray}{#1}}}}
\begin{tabular}{llccc}
\toprule
\textbf{Role} & \textbf{Model} & \textbf{Reasoning} & \textbf{Temp.} & \textbf{top-$p$} \\
\midrule
\grouphead{Discriminator panel (frozen)} \\
Discriminator & DeepSeek V4 Flash & default        & 0.0 & 1.0 \\
Discriminator & GPT-OSS-120B      & default    & 0.0 & 1.0 \\
Discriminator & Qwen3 Coder       & default            & 0.0 & 1.0 \\
Discriminator & GLM-5             & default        & 0.0 & 1.0 \\
\addlinespace
\grouphead{Generation and validation} \\
Generator     & Gemini 3.1 Pro    & default        & 0.7 & 1.0 \\
Verifier      & GPT-5.6 Luna      & medium         & default & default \\
Expert        & GPT-5.3-Codex     & default        & 0.0 & 1.0 \\
\addlinespace
\grouphead{Held-out transfer} \\
Held-out      & Kimi K3           & default        & 0.0 & 1.0 \\
\bottomrule
\end{tabular}
\end{adjustbox}
\caption{Model roles and decoding settings. \textbf{Reasoning}: native
reasoning mode.
\emph{capped} a mandatory reasoning budget excluded from the output.
\emph{default} marks parameters left to the provider.}
\label{tab:decoding}
\end{table}

% \section{Full Per-Family Results}\label{app:results}

\section{Failure Taxonomy Details}\label{app:taxonomy}

\begin{table*}[t]
\centering
\small
\begin{tabular}{l p{9.2cm} rr}
\toprule
\textbf{Failure type} & \textbf{Typical cause} & \textbf{\textsc{npc}} & \textbf{\textsc{p}} \\
\midrule
\multicolumn{4}{l}{\textit{Macro categories (\% of failing cases)}} \\
Wrong answer & Program exits cleanly, but its output is rejected by the
validator or by exact match against the reference & 33.2 & 46.6 \\
Runtime error & Program terminates with an uncaught exception and produces
no usable answer & 44.1 & 23.2 \\
Time-limit exceeded & Program is killed at the per-test wall-clock limit
& 22.7 & 30.2 \\
\midrule
\multicolumn{4}{l}{\textit{Runtime-error subtypes (\% of sampled \textsc{npc}
crashes)}} \\
\addlinespace[2pt]
\multicolumn{2}{l}{\quad Malformed code} & 48.3 & -- \\
\quad SyntaxError & Code does not parse: truncated at the token budget, or
reasoning, prose, or pseudo-code inside the code block & 42.4 & -- \\
\quad IndentationError & Same cause as SyntaxError & 4.1 & -- \\
\quad NameError & Undefined symbol, often from truncated code & 1.8 & -- \\
\addlinespace[2pt]
\multicolumn{2}{l}{\quad Input-format misread} & 27.1 & -- \\
\quad ValueError & Wrong token parsed, or wrong number of values unpacked
& 14.7 & -- \\
\quad IndexError & Off-by-one, or a structure assumed that the input does
not have & 11.2 & -- \\
\quad KeyError & Missing dictionary key from a wrong structural assumption
& 1.2 & -- \\
\addlinespace[2pt]
\multicolumn{2}{l}{\quad Scale and representation} & 12.3 & -- \\
\quad RecursionError & Deep recursion (DFS or backtracking) on large
instances without an iterative form & 8.8 & -- \\
\quad OverflowError & Floats used where large integers are required & 3.5 & -- \\
\addlinespace[2pt]
\multicolumn{2}{l}{\quad Other logic and environment errors} & 11.7 & -- \\
\quad UnboundLocalError & Variable used before assignment on some control
path & 4.7 & -- \\
\quad ModuleNotFoundError & Import of an unavailable non-standard package
(e.g.\ \texttt{numpy}, \texttt{networkx}) & 4.1 & -- \\
\quad TypeError & Mixed types or wrong argument shapes & 2.9 & -- \\
\bottomrule
\end{tabular}
\caption{Failure taxonomy. Macro categories give the share of failing test
cases per class; \textsc{npc} shares are corrected using the re-run sample.
Runtime-error subtypes are measured on the \textsc{npc} crashes only;
they sum to 99.4\%, and the remaining 0.6\% are not attributed to a listed
exception.}
\label{tab:taxonomy_full}
\end{table*}

\paragraph{Measurement.} Every graded test case is either a pass or a
failure tagged with a reason. The \textsc{p} harness records wrong answers,
runtime errors, and timeouts separately. The \textsc{npc} timing grader
records only correctness, elapsed time, and timeouts, so a program that
crashes is counted as a wrong answer. To correct for this, we re-ran a random
sample of 300 failing \textsc{npc} wrong-answer cases and captured each
process's exit status and final exception: 170 (56.7\%) were crashes, 128
(42.7\%) genuine wrong answers, and 2 (0.7\%) borderline-slow runs. We apply
these proportions to the 77.8\% of \textsc{npc} failures tagged as wrong
answers; with 300 samples, the crash share carries a 95\% confidence interval
of about $\pm$5.6 points. All figures are case-level and micro-averaged, so
they differ from the statement-level pass@1 in Table~\ref{tab:bench_results}.
Cases where a model produced no code are negligible and omitted.

\begin{table}[t]
\centering
\small
\begin{tabular}{l rr}
\toprule
\textbf{Family} & \textbf{Wrong + crash} & \textbf{TLE} \\
\midrule
01 Graph colouring      & 96.1 & 3.9 \\
02 Satisfiability      & 98.6 & 1.4 \\
03 3-D matching        & 99.4 & 0.6 \\
04 Clique / ind.\ set  & 81.6 & 18.4 \\
05 Steiner tree        & 80.6 & 19.4 \\
06 Job-shop            & 95.8 & 4.2 \\
07 Exact cover         & 53.4 & \textbf{46.6} \\
08 Linear arrangement  & 54.1 & \textbf{45.9} \\
09 Multicommodity flow & 97.4 & 2.6 \\
10 Edge colouring      & 75.9 & 24.1 \\
11 Multiway cut        & 75.7 & 24.3 \\
12 Precedence sched.   & 88.0 & 12.0 \\
\bottomrule
\end{tabular}
\caption{Share (\%) of each family's \textsc{npc} failures that are wrong
answers (including crashes) or timeouts.}
\label{tab:taxonomy_family}
\end{table}

\begin{table}[t]
\centering
\small
\setlength{\tabcolsep}{4pt}
\begin{tabular}{l rrrr}
\toprule
\textbf{Model} & \textbf{Pass} & \textbf{Wrong} & \textbf{TLE} & \textbf{Crash} \\
\midrule
Gemini-3.8-Flash    & 96.0 & 53.0 & 4.0  & 43.0 \\
Qwen3.8-Flash       & 86.3 & 53.8 & 17.8 & 28.4 \\
Qwen3.8-Omni-Flash  & 84.4 & 44.4 & 9.3  & 46.3 \\
Mistral-Medium-3.5  & 74.0 & 69.4 & 5.8  & 24.7 \\
DeepSeek-V4.1-Flash & 86.8 & 48.8 & 9.1  & 42.1 \\
Gemma-3-12B         & 22.1 & 10.0 & \textbf{69.7} & 20.3 \\
Gemma-3-27B         & 22.1 & 8.9  & \textbf{60.0} & 31.1 \\
Llama-4-Scout       & 49.2 & \textbf{82.1} & 3.4 & 14.4 \\
Qwen3.8-27B         & 87.8 & 75.0 & 13.3 & 11.8 \\
Gemma-3-4B          & 16.2 & \textbf{81.6} & 2.4 & 16.0 \\
\bottomrule
\end{tabular}
\caption{Failure signature per model on \textsc{p}, where all three types
are recorded directly. Pass is the case-level pass rate; the other columns
give the share of each model's failing cases.}
\label{tab:taxonomy_model}
\end{table}

\paragraph{Wrong answers and timeouts.} Inspection of genuine wrong answers
shows five recurring causes: misreading a constraint that hardening embeds in
ordinary prose (e.g.\ reading ``exactly one'' as ``at least one''), using a
greedy or heuristic method that passes the anchor cases but fails on corner
or worst-case instances, mishandling edge cases such as empty input or
disconnected graphs, violating the output format, and returning a solution
for an infeasible instance or vice versa. Timeouts come mainly from
exponential backtracking on worst-case instances and, less often, from
correct but unoptimised code.

\section{Qualitative Examples}
\label{app:examples}

This appendix pairs ten hardened \textsc{npc} statements with a failing solution from an evaluated model. The cases cover every failure type of the taxonomy. For each, we give the statement's disguised title, the failing test, the reference answer, and the observed outcome. The lines responsible for the failure are shown in \textcolor{red}{red}, and the explanation follows each box.

%------------------------------------------------------------
\subsection{Edge Colouring}
\label{app:ex-edge}

\noindent\textbf{Setup.} Statement \emph{Edge Assignment} (Family 10, 3-edge-colouring of a cubic graph), solved by Qwen3.8-Omni-Flash. The failing test has 8 vertices and 12 edges; the reference answer is \texttt{YES} with a valid colouring. The program raises \texttt{NameError}.

\begin{tcolorbox}[qualbox]
\begin{lstlisting}[style=qual]
import sys

def solve():
    # Read input
    # ...
    
    # Check Bipartite
    # ...
    
    @if is_bipartite:@
        # Construct Solution
        # ...
        print("YES")
        @print(...)@
    else:
        # Backtracking
        # ...
        @if bt_solve():@
            print("YES")
            print(...)
        else:
            print("NO")

if __name__ == '__main__':
    solve()
\end{lstlisting}
\end{tcolorbox}

\noindent\textbf{Why it fails.} The program is a skeleton in which every step is a placeholder comment. It never reads the input, and its first executable statement tests \texttt{is\_bipartite}, a name that is never defined, so it raises \texttt{NameError} before producing any output. Had it got further, \texttt{print(...)} would print Python's \texttt{Ellipsis} object rather than a colouring, and the fallback branch calls \texttt{bt\_solve}, which is also undefined. The outline is itself telling: the model plans to test for bipartiteness, the structure that makes the polynomial twin tractable by K\H{o}nig's theorem, and to fall back on backtracking otherwise. A strong model produced a plan built around the tractable twin and committed none of it to code.

%------------------------------------------------------------
\subsection{Steiner Tree (OverflowError)}
\label{app:ex-steiner-dp}

\noindent\textbf{Setup.} Statement \emph{Decentralized Ledger Synchronization} (Family 05, Steiner tree with a terminal subset), solved by DeepSeek-V4.1-Flash. The failing worst-case test has 180 nodes, 358 edges, and $t = 120$ terminals; its header is \texttt{180 358 120 551}, where the last value is the cost bound. The reference answer is \texttt{YES}. The program raises \texttt{OverflowError}.

\begin{tcolorbox}[qualbox]
\begin{lstlisting}[style=qual]
import sys
import heapq

def main():
    data = sys.stdin.buffer.read().split()
    if not data:
        return
    it = iter(data)
    n = int(next(it))
    m = int(next(it))
    t = int(next(it))
    @terminals = [int(next(it)) for _ in range(t)]@
    
    if t <= 1:
        print(0)
        return

    term_index = {v: i for i, v in enumerate(terminals)}
    full_mask = (1 << t) - 1

    adj = [[] for _ in range(n + 1)]
    for _ in range(m):
        u = int(next(it)); v = int(next(it)); w = int(next(it))
        adj[u].append((v, w))
        adj[v].append((u, w))

    INF = 10**18
    @dp = [[INF] * (1 << t) for _ in range(n + 1)]@

    for i, v in enumerate(terminals):
        dp[v][1 << i] = 0

    @for mask in range(1, 1 << t):@
        sub = (mask - 1) & mask
        while sub:
            other = mask ^ sub
            if sub < other:
                for v in range(1, n + 1):
                    val = dp[v][sub] + dp[v][other]
                    if val < dp[v][mask]:
                        dp[v][mask] = val
            sub = (sub - 1) & mask

        pq = []
        for v in range(1, n + 1):
            if dp[v][mask] < INF:
                heapq.heappush(pq, (dp[v][mask], v))
        while pq:
            d, u = heapq.heappop(pq)
            if d > dp[u][mask]:
                continue
            for v, w in adj[u]:
                nd = d + w
                if nd < dp[v][mask]:
                    dp[v][mask] = nd
                    heapq.heappush(pq, (nd, v))

    ans = min(dp[v][full_mask] for v in range(1, n + 1))
    @print(ans)@

if __name__ == "__main__":
    main()
\end{lstlisting}
\end{tcolorbox}

\noindent\textbf{Why it fails.} The core of the program is the Dreyfus--Wagner dynamic program over terminal subsets, a correct exact method whose table holds $2^t$ entries per vertex. With $t = 120$, the expression \texttt{[INF] * (1 << t)} requests a list of $2^{120}$ elements, a length Python cannot even represent as an index, so the allocation raises \texttt{OverflowError} before any computation starts; the subset loop that follows would run for $2^{120}$ iterations in any case. The program also misreads the input contract. The header carries a fourth integer, the cost bound, and the task is to decide whether a tree within that bound exists. Reading only three header fields, the program consumes the bound as the first terminal and shifts every later value by one, and it prints a cost where a \texttt{YES}/\texttt{NO} verdict is expected. It would therefore fail even on an instance with few terminals: the case shows both effects of hardening at once, a disguised I/O contract and a parameter pushed past the range where the exact method is feasible.

%------------------------------------------------------------
\subsection{Linear Arrangement}
\label{app:ex-linarr}

\noindent\textbf{Setup.} Statement \emph{Workbench Layout} (Family 08, minimum linear arrangement of a general graph), solved by DeepSeek-V4.1-Flash. The failing test has $n = 50$ vertices and 113 edges; the reference answer is \texttt{YES} with a valid arrangement. The program raises \texttt{MemoryError}.

\begin{tcolorbox}[qualbox]
\begin{lstlisting}[style=qual]
import sys

def solve():
    data = sys.stdin.read().split()
    if not data:
        return
    it = iter(data)
    n = int(next(it))
    m = int(next(it))
    B = int(next(it))
    adj = [0] * n
    for _ in range(m):
        u = int(next(it)) - 1
        v = int(next(it)) - 1
        adj[u] |= (1 << v)
        adj[v] |= (1 << u)
    deg = [adj[i].bit_count() for i in range(n)]
    @size = 1 << n@
    @cut = [0] * size@
    for mask in range(1, size):
        v = (mask & -mask).bit_length() - 1
        prev = mask ^ (1 << v)
        cut[mask] = cut[prev] + deg[v] - 2 * (adj[v] & prev).bit_count()
    INF = 10**9
    @dp = [INF] * size@
    dp[0] = 0
    @parent = [-1] * size@
    for mask in range(1, size):
        m = mask
        best = INF
        best_v = -1
        while m:
            v = (m & -m).bit_length() - 1
            m &= m - 1
            prev = mask ^ (1 << v)
            cost = dp[prev] + cut[mask]
            if cost < best:
                best = cost
                best_v = v
        dp[mask] = best
        parent[mask] = best_v
    full_mask = size - 1
    if dp[full_mask] > B:
        print("NO")
    else:
        print("YES")
        perm = []
        mask = full_mask
        while mask:
            v = parent[mask]
            perm.append(v)
            mask ^= (1 << v)
        perm.reverse()
        pos = [0] * n
        for i, op in enumerate(perm):
            pos[op] = i + 1
        print(' '.join(map(str, pos)))

if __name__ == "__main__":
    solve()
\end{lstlisting}
\end{tcolorbox}

\noindent\textbf{Why it fails.} The method is sound. The cost of an arrangement equals the sum, over every prefix of the ordering, of the number of edges leaving that prefix, so a table indexed by vertex subsets, with \texttt{cut[mask]} as the prefix term, finds the optimum exactly. Unlike the previous example, the program also reads the bound and answers in the required format. The table, however, has one entry per subset of the vertices, and with $n = 50$ the first allocation, \texttt{cut = [0] * size}, requests about $1.1 \times 10^{15}$ entries and raises \texttt{MemoryError}; \texttt{dp} and \texttt{parent} would need two more arrays of the same size. As in Appendix~\ref{app:ex-steiner-dp}, a correct exponential method is defeated by the instance size rather than by faulty reasoning, here on the memory axis.

%------------------------------------------------------------
\subsection{Exact Cover}
\label{app:ex-x3c}

\noindent\textbf{Setup.} Statement \emph{Full-Coverage Cell Deployment} (Family 07, exact cover by 3-sets), solved by DeepSeek-V4.1-Flash. The failing test has $q = 55$, hence 165 elements, and 165 candidate triples; the reference answer is \texttt{YES} with a partition into 55 triples. The program exceeds the time limit.

\begin{tcolorbox}[qualbox]
\begin{lstlisting}[style=qual]
import sys
sys.setrecursionlimit(10000)

def solve():
    data = sys.stdin.read().strip().split()
    if not data:
        return
    q = int(data[0])
    t = int(data[1])
    n = 3 * q
    triples = []
    idx = 2
    for _ in range(t):
        a = int(data[idx]) - 1
        b = int(data[idx+1]) - 1
        c = int(data[idx+2]) - 1
        idx += 3
        triples.append((a, b, c))
    
    triple_masks = []
    for a, b, c in triples:
        mask = (1 << a) | (1 << b) | (1 << c)
        triple_masks.append(mask)
    
    elem_triples = [[] for _ in range(n)]
    for i, (a, b, c) in enumerate(triples):
        elem_triples[a].append(i)
        elem_triples[b].append(i)
        elem_triples[c].append(i)
    
    full_mask = (1 << n) - 1
    selected = []
    
    @def dfs(covered):@
        if covered == full_mask:
            return True
        uncovered = (~covered) & full_mask
        @i = (uncovered & -uncovered).bit_length() - 1@
        @for ti in elem_triples[i]:@
            mask = triple_masks[ti]
            if mask & covered == 0:
                selected.append(ti)
                if dfs(covered | mask):
                    return True
                selected.pop()
        return False
    
    if dfs(0):
        print("YES")
        for ti in selected:
            a, b, c = triples[ti]
            print(a+1, b+1, c+1)
    else:
        print("NO")

if __name__ == "__main__":
    solve()
\end{lstlisting}
\end{tcolorbox}

\noindent\textbf{Why it fails.} This is a correct exact-cover search: it picks an uncovered element and tries every disjoint triple that contains it, so any answer it returns is right. What it lacks is a rule that keeps the search small. It always branches on the lowest-numbered uncovered element rather than on the element with the fewest remaining candidate triples, the choice at the heart of Knuth's Algorithm~X, and it never prunes a branch in which some uncovered element can no longer be covered. The search tree therefore grows exponentially in $q$, and with 55 triples to place on a near-threshold instance the program runs past the time limit without ever printing an answer. Exact cover is one of the two families in which timeouts account for nearly half of all \textsc{npc} failures, and this program shows why: the failure is not a wrong answer but one that never arrives.

%------------------------------------------------------------
\subsection{Steiner Tree (ValueError)}
\label{app:ex-steiner-header}

\noindent\textbf{Setup.} Statement \emph{Synthetic Metabolism Design} (Family 05, Steiner tree), solved by Gemma-3-27B on the same worst-case test as Appendix~\ref{app:ex-steiner-dp}: the header \texttt{180 358 120 551} gives the node count, edge count, terminal count, and cost bound. The reference answer is \texttt{YES}. The program raises \texttt{ValueError}.

\begin{tcolorbox}[qualbox]
\begin{lstlisting}[style=qual]
def solve():
    @n, m, t = map(int, input().split())@
    vital = list(map(int, input().split()))
    edges = []
    for _ in range(m):
        u, v, w = map(int, input().split())
        edges.append((u, v, w))

    if t == 1:
        print(0)
        return

    adj = {i: [] for i in range(1, n + 1)}
    for u, v, w in edges:
        adj[u].append((v, w))
        adj[v].append((u, w))

    min_cost = float('inf')

    @for i in range(1 << m):@
        selected_edges = []
        cost = 0
        for j in range(m):
            if (i >> j) & 1:
                selected_edges.append(edges[j])
                cost += edges[j][2]

        selected_adj = {k: [] for k in range(1, n + 1)}
        for u, v, w in selected_edges:
            selected_adj[u].append((v, w))
            selected_adj[v].append((u, w))

        def bfs(start_node, end_node):
            q = [(start_node, 0)]
            visited = {start_node}
            while q:
                node, dist = q.pop(0)
                if node == end_node:
                    return True
                for neighbor, _ in selected_adj[node]:
                    if neighbor not in visited:
                        visited.add(neighbor)
                        q.append((neighbor, dist + 1))
            return False

        connected = True
        for start in vital:
            for end in vital:
                if start != end and not bfs(start, end):
                    connected = False
                    break
            if not connected:
                break

        if connected:
            min_cost = min(min_cost, cost)

    @print(min_cost)@

solve()
\end{lstlisting}
\end{tcolorbox}

\noindent\textbf{Why it fails.} The first line unpacks three integers from a header that holds four. The hardened statement introduces the cost bound in prose rather than as a labelled input field, and the model never registers it as part of the input, so the unpacking raises \texttt{ValueError} and nothing else runs. The rest of the program could not have rescued it. It enumerates every subset of the $m = 358$ edges, $2^{358}$ iterations in total, tests connectivity between every pair of terminals with a breadth-first search that pops from the front of a list, and prints a minimum cost where the task asks for a \texttt{YES}/\texttt{NO} verdict. The crash on the header hides a solution that was hopeless on every other count, a pattern typical of the input-format misreads that make up over a quarter of \textsc{npc} crashes.

%------------------------------------------------------------
\subsection{Multicommodity Flow}
\label{app:ex-mcf}

\noindent\textbf{Setup.} Statement \emph{Intermodal Rail Network Routing} (Family 09, integral multicommodity flow), solved by Llama-4-Scout. The failing test has 8 nodes, 14 edges, and 3 commodities; the reference answer is \texttt{YES} with an integral flow for every commodity on every edge. The program raises \texttt{ModuleNotFoundError}.

\begin{tcolorbox}[qualbox]
\begin{lstlisting}[style=qual]
import numpy as np
@from scipy.optimize import linprog@

def solve():
    n, m, k = map(int, input().split())
    contracts = []
    for _ in range(k):
        s, t, d = map(int, input().split())
        contracts.append((s-1, t-1, d))

    graph = [[] for _ in range(n)]
    limits = []
    for i in range(m):
        u, v, c = map(int, input().split())
        graph[u-1].append((v-1, i))
        limits.append(c)

    A_eq = []
    b_eq = []
    for contract in contracts:
        s, t, d = contract
        row = [0] * (k * m)
        for j in range(m):
            @row[j * k + contracts.index(contract)] = 1@
        A_eq.append(row)
        b_eq.append(d)

    A_ub = []
    b_ub = []
    for i in range(m):
        row = [0] * (k * m)
        for j in range(k):
            row[i * k + j] = 1
        A_ub.append(row)
        b_ub.append(limits[i])

    c = [0] * (k * m)
    bounds = [(0, None)] * (k * m)

    try:
        @res = linprog(c, A_ub=A_ub, b_ub=b_ub, A_eq=A_eq, b_eq=b_eq, bounds=bounds, method="highs")@
        if res.success:
            print("YES")
            for i in range(k):
                row = []
                for j in range(m):
                    @row.append(int(res.x[i * m + j]))@
                print(*row)
        else:
            print("NO")
    except ValueError:
        print("NO")

solve()
\end{lstlisting}
\end{tcolorbox}

\noindent\textbf{Why it fails.} The immediate failure is the import: SciPy is not installed in the evaluation environment, so the program raises \texttt{ModuleNotFoundError} before reading any input. The model behind the import is also wrong. Each commodity's equality row adds up its flow over every edge in the network and sets that total equal to the demand, and no constraint enforces conservation at any node; the adjacency list \texttt{graph} is built and never used, so a feasible solution of this program need not be a flow at all. Even with correct constraints, \texttt{linprog} solves the continuous relaxation, which is polynomial and may return fractional flows, while the task requires an integral witness; truncating with \texttt{int()} can then break both demands and conservation. Finally, variables are laid out as \texttt{j * k + i} but read back as \texttt{i * m + j}, so the printed flows would be attributed to the wrong commodities and edges. This is the characteristic failure that Appendix~B anticipates for this family: the model solves the tractable linear relaxation instead of the integral problem.

%------------------------------------------------------------
\subsection{Satisfiability}
\label{app:ex-sat}

\noindent\textbf{Setup.} Statement \emph{Control Panel Commissioning} (Family 02, 3-SAT), solved by Llama-4-Scout. The failing corner test is the input \texttt{1 0}: one variable and no clauses, which is trivially satisfiable. The reference answer is \texttt{YES} with any assignment. The program raises \texttt{TypeError}.

\begin{tcolorbox}[qualbox]
\begin{lstlisting}[style=qual]
import sys
from array import array

def read_input():
    n, m = map(int, sys.stdin.buffer.readline().split())
    interlocks = []
    for _ in range(m):
        l1, l2, l3 = map(int, sys.stdin.buffer.readline().split())
        interlocks.append((l1, l2, l3))
    return n, m, interlocks

def solve(n, interlocks):
    for i in range(1 << n):
        is_valid = True
        for l1, l2, l3 in interlocks:
            if (l1 > 0 and (i >> (l1 - 1) & 1) == 0) or (l1 < 0 and (i >> (-l1 - 1) & 1) == 1):
                pass
            elif (l2 > 0 and (i >> (l2 - 1) & 1) == 0) or (l2 < 0 and (i >> (-l2 - 1) & 1) == 1):
                pass
            elif (l3 > 0 and (i >> (l3 - 1) & 1) == 0) or (l3 < 0 and (i >> (-l3 - 1) & 1) == 1):
                pass
            @else:@
                @is_valid = False@
                break
        if is_valid:
            return i
    return -1

def print_output(result):
    if result == -1:
        print('NO')
    else:
        print('YES')
        @print(' '.join(array('i', [(result >> i) & 1 for i in range(n)])))@

n, _, interlocks = read_input()
result = solve(n, interlocks)
print_output(result)
\end{lstlisting}
\end{tcolorbox}

\noindent\textbf{Why it fails.} The program enumerates assignments correctly, but its clause test is inverted. Each branch of the \texttt{if}/\texttt{elif} chain checks whether a literal is \emph{false} and, if so, accepts the clause, so a clause counts as satisfied as soon as any one of its literals is false, and the assignment is rejected only when all three literals are true. The correct rule is the opposite: a clause fails exactly when all three literals are false. On this test the bug stays hidden, because with no clauses the first assignment is accepted and the program prints \texttt{YES}. It then crashes while printing the assignment: \texttt{array('i', \ldots)} holds integers, and \texttt{str.join} accepts only strings, so the call raises \texttt{TypeError}. The easiest instance in the suite is lost to an output-formatting error, before the inverted logic ever has a chance to matter.

%------------------------------------------------------------
\subsection{Precedence Scheduling}
\label{app:ex-prec}

\noindent\textbf{Setup.} Statement \emph{Culinary Event Sequencing} (Family 12, precedence-constrained scheduling on several machines), solved by Gemma-3-12B. The failing anchor test has the header \texttt{6 3 5 9}: six tasks, three stations, a horizon of five time slots, and nine precedence pairs. The reference answer is \texttt{YES} with the schedule \texttt{1 1 2 3 4 5}. The program raises \texttt{IndexError} on this first anchor test.

\begin{tcolorbox}[qualbox]
\begin{lstlisting}[style=qual]
def solve():
    n, m, t, k = map(int, input().split())
    edges = []
    for _ in range(k):
        u, v = map(int, input().split())
        edges.append((u, v))

    def check(assignment):
        for u, v in edges:
            if assignment[u-1] >= assignment[v-1]:
                return False
        return True

    def is_valid(assignment):
        for i in range(t):
            count = 0
            for j in range(n):
                if assignment[j] == i + 1:
                    count += 1
            if count > m:
                return False
        return True

    import itertools
    @for perm in itertools.permutations(range(1, t + 1)):@
        assignment = [0] * n
        for i in range(n):
            @assignment[i] = perm[i]@
        
        if check(assignment) and is_valid(assignment):
            print("YES")
            print(*assignment)
            return

    print("NO")

solve()
\end{lstlisting}
\end{tcolorbox}

\noindent\textbf{Why it fails.} The program mixes up the number of tasks with the number of time slots. It enumerates permutations of the five slot values and then assigns \texttt{perm[i]} to each of the six tasks, so the sixth assignment indexes past the end of a tuple of length five and raises \texttt{IndexError}. The indexing error hides a deeper modelling error. A permutation gives every task a different slot, but the point of having several stations is that tasks can share a slot, as the reference schedule does by placing the first two tasks in slot~1. The capacity check in \texttt{is\_valid} shows the model knew that stations limit how many tasks share a slot, yet its search never generates a schedule in which any slot is shared. With more tasks than slots, no permutation can give each task its own slot, so even with correct indexing the program could never answer \texttt{YES}.

%------------------------------------------------------------
\subsection{Job-Shop Scheduling (Corner Case 1)}
\label{app:ex-jobshop-seq}

\noindent\textbf{Setup.} Statement \emph{Facility Coordination} (Family 06, job-shop makespan), solved by Mistral-Medium-3.5. The failing corner test has two jobs on two machines, each consisting of two 10-unit operations that visit the machines in opposite orders. The reference optimum is \texttt{20}; the program outputs \texttt{40}.

\begin{tcolorbox}[qualbox]
\begin{lstlisting}[style=qual]
import sys
from itertools import permutations

def main():
    data = sys.stdin.read().split()
    if not data:
        return
    
    j = int(data[0])
    m = int(data[1])
    orders = []
    index = 2
    for _ in range(j):
        o = int(data[index])
        index += 1
        phases = []
        for __ in range(o):
            unit = int(data[index])
            duration = int(data[index+1])
            phases.append((unit, duration))
            index += 2
        orders.append(phases)
    
    units = list(range(1, m+1))
    min_time = float('inf')
    
    @for perm in permutations(units):@
        unit_order = {u: i for i, u in enumerate(perm)}
        end_times = [0] * m
        order_finish = [0] * j
        
        @for order_idx in range(j):@
            current_time = 0
            for unit, duration in orders[order_idx]:
                u_idx = unit_order[unit]
                @start_time = max(current_time, end_times[u_idx])@
                end_time = start_time + duration
                end_times[u_idx] = end_time
                current_time = end_time
            order_finish[order_idx] = current_time
        
        max_finish = max(order_finish)
        if max_finish < min_time:
            min_time = max_finish
    
    print(min_time)

if __name__ == "__main__":
    main()
\end{lstlisting}
\end{tcolorbox}

\noindent\textbf{Why it fails.} Running the first job on machine~1 while the second runs on machine~2, and then swapping, finishes both jobs in 20 units. The program cannot find this schedule. It places the jobs greedily in input order, starting each operation after the latest finish already booked on its machine, so the second job's first operation waits until the first job has released machine~2 and the makespan comes out at 40. The outer loop over machine permutations appears intended to explore alternatives, but it only relabels machine indices and leaves every computed time unchanged, so the program never considers a different job order or any interleaving of the jobs. The output is a feasible schedule, just not an optimal one, and only exact-match grading against the reference optimum exposes the error.

%------------------------------------------------------------
\subsection{Job-Shop Scheduling (Corner Case 2)}
\label{app:ex-jobshop-load}

\noindent\textbf{Setup.} Statement \emph{Linear Pipeline Interleaving} (Family 06, job-shop makespan), solved by Mistral-Medium-3.5. The failing corner test has a single job of ten unit-length operations, five on each of two machines, which must run in sequence. The reference optimum is \texttt{10}; the program outputs \texttt{5}.

\begin{tcolorbox}[qualbox]
\begin{lstlisting}[style=qual]
import sys
from itertools import permutations

def main():
    data = sys.stdin.read().split()
    if not data:
        return
    
    j = int(data[0])
    m = int(data[1])
    jobs = []
    index = 2
    for _ in range(j):
        o = int(data[index])
        index += 1
        ops = []
        for __ in range(o):
            machine = int(data[index])
            duration = int(data[index+1])
            ops.append((machine, duration))
            index += 2
        jobs.append(ops)
    
    total_ops = sum(len(job) for job in jobs)
    if total_ops == 0:
        print(0)
        return
    
    machines = list(range(1, m+1))
    best = float('inf')
    
    for machine_order in permutations(machines):
        machine_order = list(machine_order)
        machine_to_pos = {m: i for i, m in enumerate(machine_order)}
        
        schedule = [[] for _ in range(m)]
        for job in jobs:
            for machine, duration in job:
                pos = machine_to_pos[machine]
                @schedule[pos].append(duration)@
        
        max_time = 0
        for machine_schedule in schedule:
            current_time = 0
            for duration in machine_schedule:
                current_time += duration
            if current_time > max_time:
                @max_time = current_time@
        if max_time < best:
            best = max_time
    
    print(best)

if __name__ == "__main__":
    main()
\end{lstlisting}
\end{tcolorbox}

\begin{figure*}[!t]
\centering
\includegraphics[width=\textwidth]{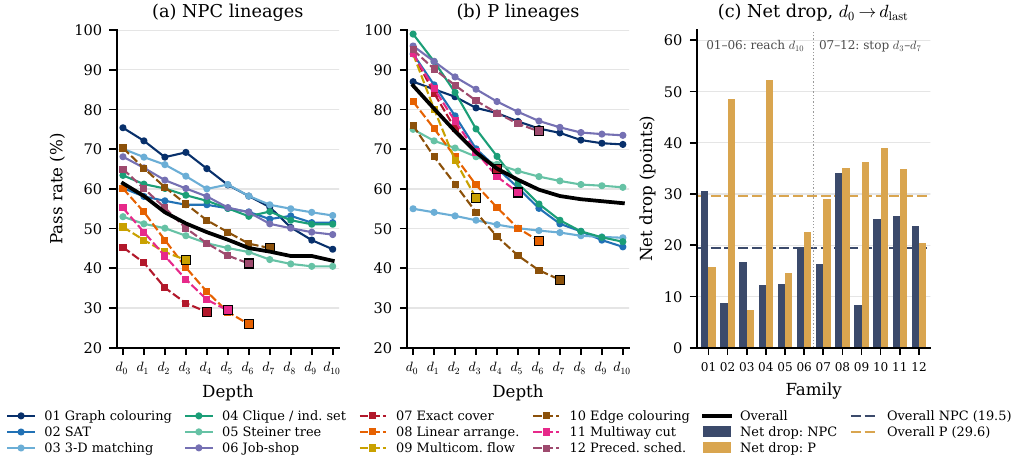}
\caption{Accuracy along the hardening ladder for \textsc{npc} and \textsc{p} lineages. \textbf{(a, b)}~Evaluation-panel pass@1\% accuracy at each depth for \textsc{npc} lineages (a) and \textsc{p} lineages (b), on a common $y$-axis. Families 01--06 (solid lines, circles) reach $d_{10}$; families 07--12 (dashed lines, squares) terminate between $d_3$ and $d_7$, and an outlined square marks the last depth each reaches. The thick black curve is the overall pass rate for the class. \textbf{(c)}~Net drop from $d_0$ to the last depth reached, in percentage points, for each family and class; dashed horizontal lines mark the overall net drops.}
\label{fig:depth_drop}
\end{figure*}

\noindent\textbf{Why it fails.} The program discards the order of operations within a job. It files each operation's duration under its machine and then takes the makespan to be the largest machine load, which here is 5. That quantity is a lower bound on the makespan, not the makespan itself: it ignores that each operation must wait for its predecessor in the same job, and because the ten operations of this job must run one after another, the true makespan is 10. The permutation loop is again inert. Together with Appendix~\ref{app:ex-jobshop-seq}, this example shows the same model misreading what the makespan measures in opposite directions, once by serialising jobs that could overlap and once by overlapping operations that must be serial. Both errors produce plausible numbers, and exact-match grading on the optimisation families is what catches them.

\section{Accuracy Drop Across Depths}
\label{app:depth_drop}

Figure~\ref{fig:depth_drop} shows what happens to accuracy as we keep hardening the same instances. Both classes get harder with every step. \textsc{npc} lineages fall from 61.4\% to 41.9\% (19.5 points), and \textsc{p} lineages fall from 86.0\% to 56.4\% (29.6 points). So \textsc{p} problems start out much easier but lose more, and the gap between the two classes shrinks from 24.6 to 14.5 points. Most of the damage happens early: about half of the total drop comes in the first three steps, and the last three steps barely move the curve. The decline is also steady. Every \textsc{p} step is a drop, and the few \textsc{npc} increases are at most 1.2 points, small enough to be noise.

Families do not all react the same way. Families 07--12 stop earlier on the ladder but lose accuracy two to three times faster per step than families 01--06, in both classes. The same family can also behave very differently depending on its class: graph colouring loses far more as \textsc{npc} (30.6 vs.\ 15.8 points), while SAT loses far more as \textsc{p} (48.6 vs.\ 8.7). In SAT, clique, and edge colouring, the \textsc{p} curve ends up below the \textsc{npc} one, and in 3-D matching it is below at every depth. In other words, enough hardening can make an easy problem harder than its \textsc{npc} counterpart.

\end{document}